\documentclass[10pt,journal,compsoc]{IEEEtran}

\ifCLASSOPTIONcompsoc
    \usepackage[noadjust]{cite}
\else
  \usepackage[noadjust]{cite}
\fi
\ifCLASSINFOpdf  
\else  
\fi
\usepackage{amsmath}
\usepackage{url}
\usepackage{xcolor}
\usepackage{times}
\usepackage{epsfig}
\usepackage{graphicx}
\usepackage{caption}
\usepackage{float}
\usepackage{amssymb} 
\usepackage{footmisc}
\usepackage{lineno}
\usepackage{color}
\usepackage{subcaption}
\usepackage{multirow}
\usepackage{dsfont}
\usepackage{mathtools}
\usepackage{setspace}
\usepackage{csquotes}
\usepackage{breqn}
\usepackage{lettrine}
\usepackage{upquote}
\usepackage{adjustbox}
\usepackage{wrapfig}
\usepackage{lipsum}
\usepackage{mathtools}
\usepackage{comment}
\usepackage{bm}
\usepackage{array}
\usepackage{tabu}
\usepackage{booktabs}
\usepackage{enumerate}
\DeclareCaptionFormat{myformat}{\fontsize{8}{10}\selectfont#1#2#3}
\DeclareMathAlphabet\mathbfcal{OMS}{cmsy}{b}{n}

\DeclareMathAlphabet{\pazocal}{OMS}{zplm}{m}{n}
\DeclareMathAlphabet{\mathpzc}{OT1}{pzc}{m}{it}

\usepackage[ruled,vlined]{algorithm2e}
\usepackage[colorlinks=true,linkcolor=blue]{hyperref}
\def\arrvline{\hfil\kern\arraycolsep\vline\kern-\arraycolsep\hfilneg}

\ifCLASSINFOpdf
\else
\fi

\begin{document}
\bstctlcite{IEEEexample:BSTcontrol}
\title{Predicting the Best of \textit{N} Visual Trackers}

\author{Basit Alawode,
Sajid Javed,
Arif Mahmood, 
~Jiri Matas, \IEEEmembership{CS Member,~IEEE} 
\IEEEcompsocitemizethanks{\IEEEcompsocthanksitem  B. \ Alawode, and S.\ Javed are with the Department of Computer Science, Khalifa University of Science and Technology, P.O Box: 127788, Abu Dhabi, UAE. (email: sajid.javed@ku.ac.ae).
\IEEEcompsocthanksitem A. \ Mahmood is with the Department of Computer Science, Information Technology University, Pakistan (email: arif.mahmood@itu.edu.pk).
\IEEEcompsocthanksitem J.\ Matas is with Center for Machine Perception, Czech Technical University, Prague.}}

\markboth{Journal of \LaTeX\ Class Files,~Vol.~14, No.~8, August~2015}
{Javed \etal: VOT}
\IEEEtitleabstractindextext{%


\begin{abstract}
We observe that the performance of SOTA visual trackers surprisingly strongly varies across different video attributes and datasets.
No single tracker remains the best performer across all tracking attributes and datasets.
To bridge this gap, for a given video sequence, we predict the ``Best of the $N$ Trackers'', called the BofN meta-tracker.
At its core, a Tracking Performance Prediction Network (TP$^2$N) selects a predicted best performing visual tracker for the given video sequence using only a few initial frames.
%
%
We also introduce a frame-level BofN meta-tracker which keeps predicting best performer after regular temporal intervals. 
The TP$^2$N is based on self-supervised learning architectures MocoV2, SwAv, BT, and DINO; experiments show that the DINO with ViT-S as a backbone performs the best.
The video-level BofN meta-tracker outperforms, by a large margin, existing SOTA trackers on nine standard benchmarks -- LaSOT, TrackingNet, GOT-10K, VOT2019, VOT2021, VOT2022, UAV123, OTB100, and WebUAV-3M. 
Further  improvement is achieved by the frame-level BofN meta-tracker effectively handling  variations in the tracking scenarios within long sequences.
For instance, on GOT-10k, BofN meta-tracker average overlap is 88.7$\%$ and 91.1$\%$ with video and frame-level settings respectively.
The best performing tracker, RTS, achieves 85.20$\%$ AO.
On VOT2022, BofN expected average overlap is 67.88$\%$ and 70.98$\%$ with video and frame level settings, compared to the best performing ARTrack, 64.12$\%$.
This work also presents an extensive evaluation of competitive tracking methods on all commonly used benchmarks, following their protocols.\\
The code, the trained models, and the results will soon be made publicly available on \href{https://github.com/BasitAlawode/Best\_of\_N\_Trackers}{https://github.com/BasitAlawode/Best\_of\_N\_Trackers}.
 
\end{abstract}

}

\maketitle

\IEEEdisplaynontitleabstractindextext

\IEEEpeerreviewmaketitle

\section{Introduction}
\label{sec:intro}
Visual Object Tracking (VOT) is a fundamental and active research area in computer vision \cite{9913708}.
Given the location of a generic moving target object in the first frame, the main aim of VOT is to estimate its location in the remaining video sequence \cite{8954084}.
VOT has been employed in a wide range of real-world applications, including video surveillance \cite{7428948}, robotics navigation \cite{9316980}, medical video analysis \cite{ben2022graph}, autonomous driving \cite{9354012}, animal behavior analysis \cite{martinez2014animal}, and human activity recognition \cite{beddiar2020vision}. 
VOT is a challenging problem because it requires learning a class-agnostic model of generic target objects in the presence of noise, occlusion, motion blur, and fast motion \cite{9666461}.
In addition, the target may undergo significant scale variations, and out-of-plane rotations, and be affected by illumination variations, and background-cluttered \cite{8954084}.

In recent years, VOT research has significantly progressed, 
and several learning-based visual tracking paradigms have been proposed \cite{Lukezic_2017_CVPR, 9913708, Valmadre_2017_CVPR,thangavel2023transformers}
such as the discriminative correlation filters \cite{Valmadre_2017_CVPR}, deep Siamese networks \cite{bertinetto2016fully}, transformer-driven trackers \cite{thangavel2023transformers}, and large language model-driven trackers \cite{li2023citetracker}.
In most areas, over time, the best-performing computer vision methods all tend to follow a single paradigm, such as the visual transformer in object detection \cite{10.1007/978-3-030-58452-8_13} and the paradigm becomes the de facto standard. 

\begin{figure*}[t!]
\centering
\begin{subfigure}{0.495\linewidth}
\includegraphics[width=\linewidth]{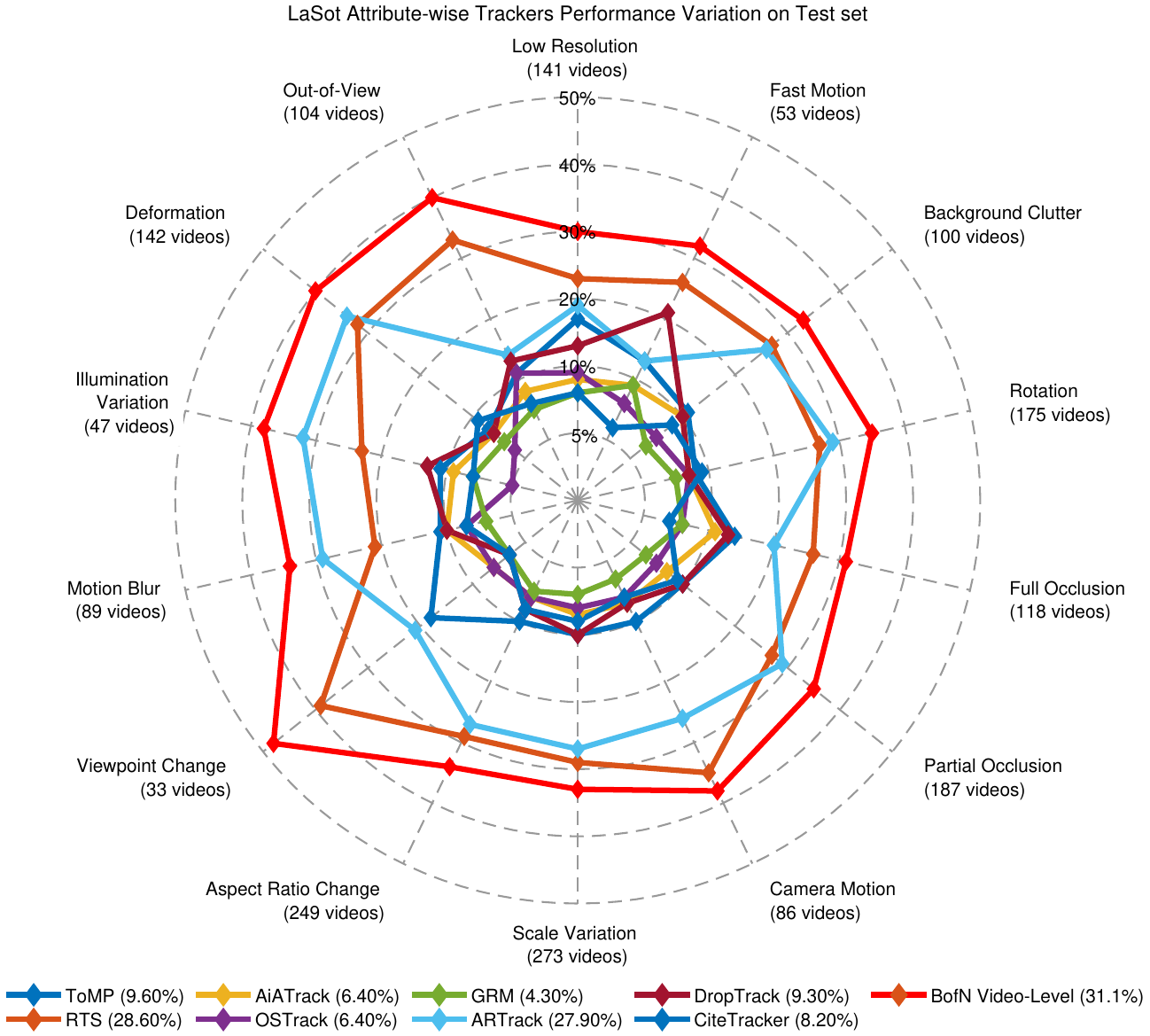}
\caption{LaSoT dataset}
\label{fig:first}
\end{subfigure}
\begin{subfigure}{0.495\linewidth}
    \includegraphics[width=\linewidth]{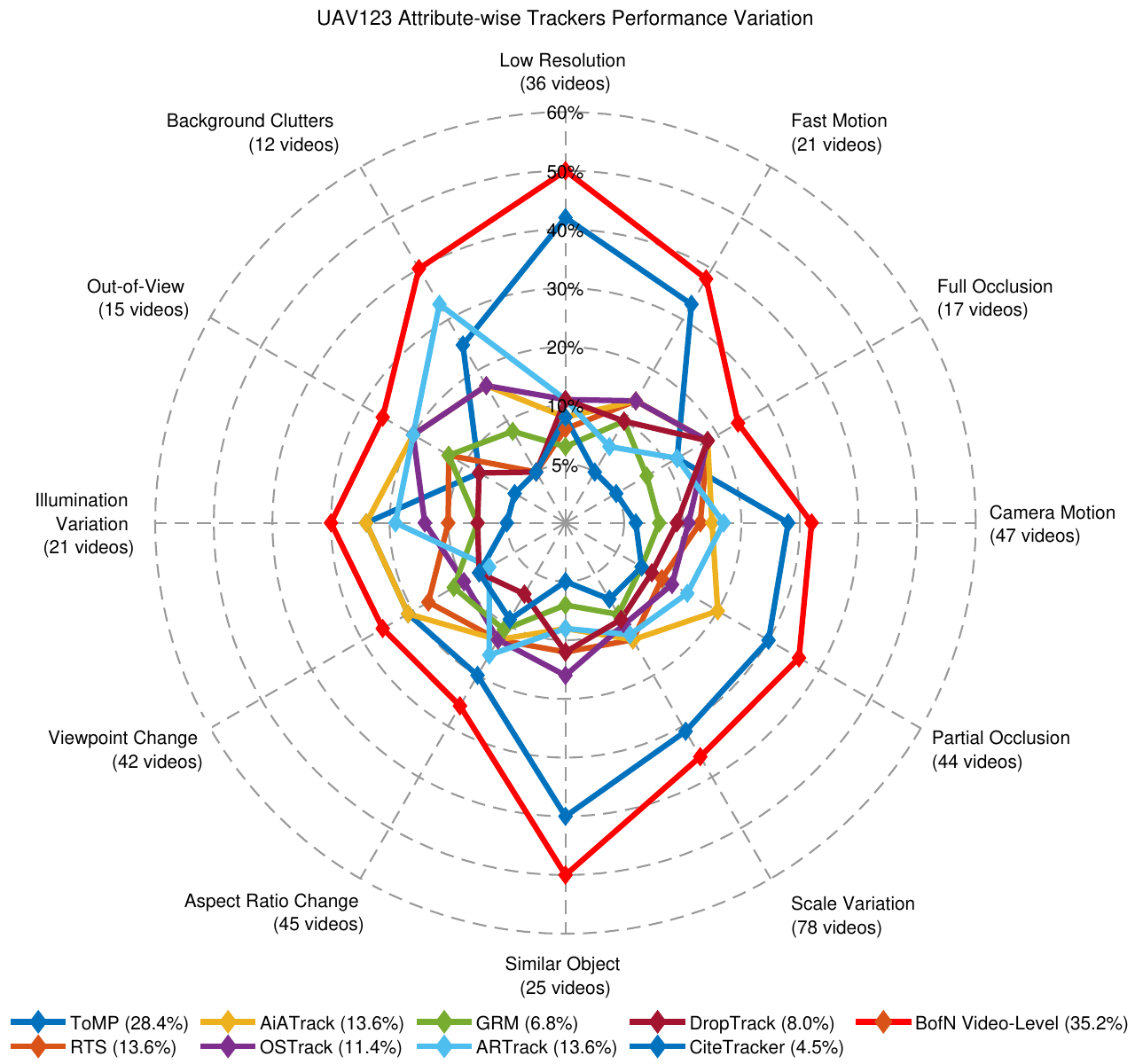}
    \caption{UAV123 dataset.}
    \label{fig:second}
\end{subfigure}
\begin{subfigure}{0.30\linewidth}
    \includegraphics[width=\linewidth]{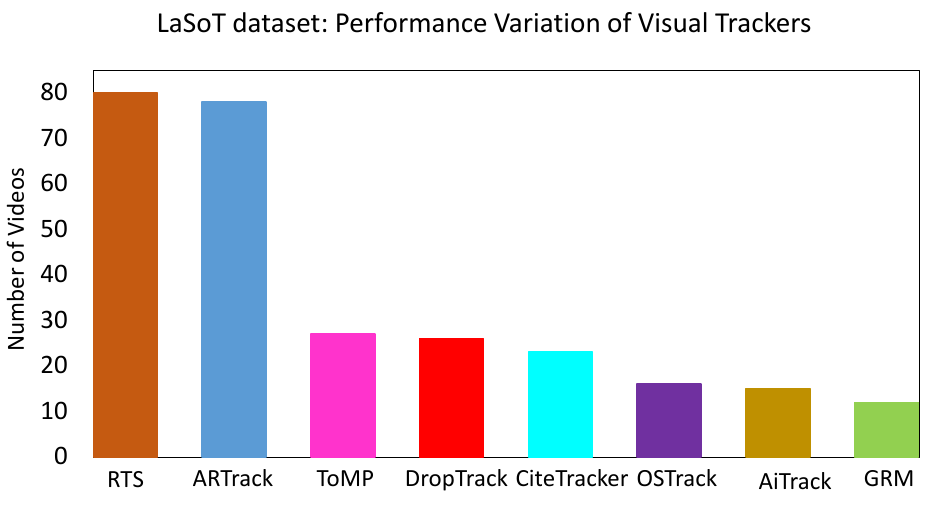}
    \caption{LaSoT dataset.}
    \label{fig:second}
\end{subfigure}
\begin{subfigure}{0.30\linewidth}
    \includegraphics[width=\linewidth]{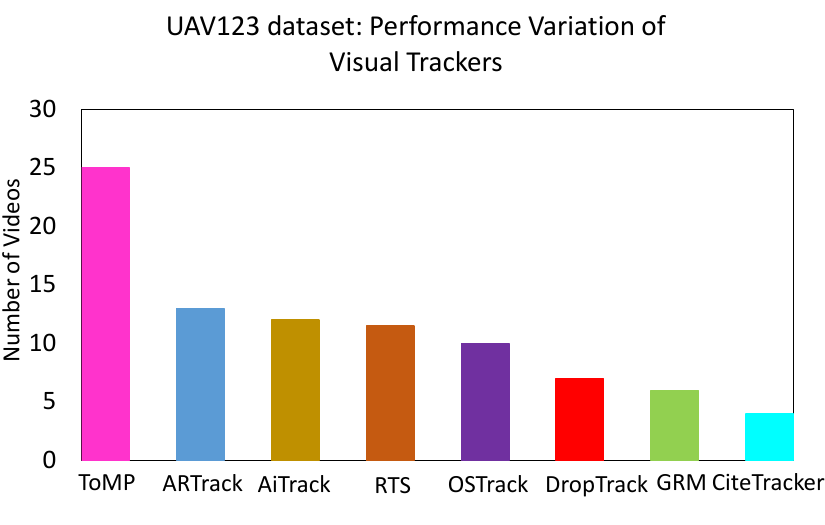}
    \caption{UAV123 dataset.}
    \label{fig:second}
\end{subfigure}
\begin{subfigure}{0.30\linewidth}
    \includegraphics[width=\linewidth]{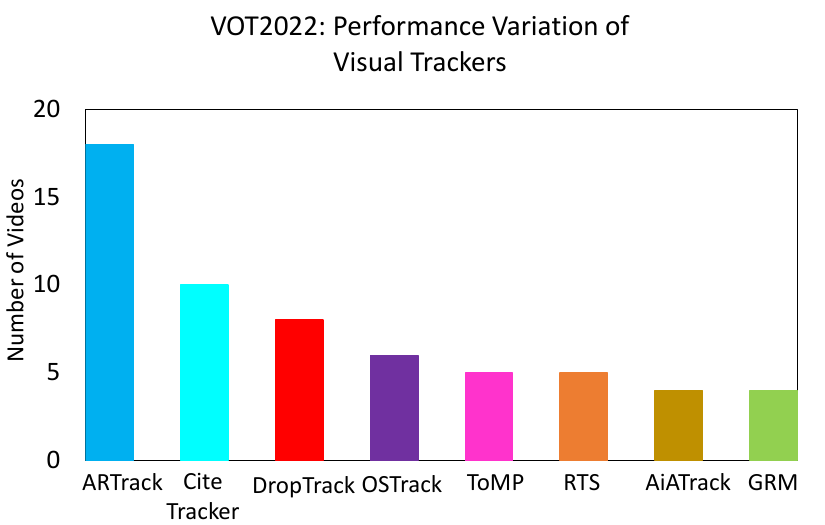}
    \caption{VOT2022 dataset.}
    \label{fig:second}
\end{subfigure}
\caption{SOTA visual trackers -- performance variation. Dependence on video attributes on (a) LaSoT  and (b) UAV123. The proposed BofN dominates for all attributes on both datasets. The umber of videos where a tracker is the top performer on (c) LaSoT, (d) UAV123 and (e) VOT2022 datasets. }
\label{fig1}
\end{figure*}

The observation that in visual object tracking, no paradigm or particular tracker currently dominates all others, is the first contribution of the paper. 
This is highlighted in Fig.~\ref{fig1}, where the performance of eight diverse SOTA trackers on three common datasets, LASOT \cite{8954084}, UAV123 \cite{10.1007/978-3-319-46448-0_27}, and VOT2022 \cite{kristan2022tenth} is shown.
We first note that, see the percentages after tracker labels at the bottom of Figs.~\ref{fig1} (a) and (b), that all the selected trackers are the best performing on at least 4.0\% of the sequences. 
Also, no tracker performs the best for all of the 14 attributes of LASOT.
For example, for the ``view change'' tracking attribute, RTS \cite{10.1007/978-3-031-20047-2_33} is the best for 38.0$\%$, ARTrack \cite{Wei_2023_CVPR} for 22.0$\%$, and ToMP \cite{mayer2022transforming}  for 18.0$\%$ sequences, respectively. 
Yet, for the ``Illumination Variation'' attribute, the ARTrack is top performing overall, followed by RTS and DropTrack.

%
%
%

To exploit the observation that tracker performance varies, one needs to predict which tracker is likely to perform best for a given sequence or a given frame.
As a second contribution to the paper, we show that reliable predictions are indeed possible and we propose a meta-tracker, called ``Best of the $N$ Trackers'' (BofN) meta-tracker,  that selects an output tracker based on tracking performance prediction network. 
The meta-tracker significantly outperforms each individual tracker.
For this purpose, a Tracking Performance Prediction Network (TP$^{2}$N) is trained in an end-to-end manner using the performance of the SOTA visual trackers as labels on a wide range of tracking scenarios.
The TP$^2$N network reliably discriminates between different tracking scenarios (see Tables \ref{table1} and \ref{table2}).

To train TP$^{2}$N, we present each training video to the \textit{N} SOTA trackers and obtain their performance scores.
Based on the best score, each video is labeled to map the tracker selection problem to the video classification task, such that for each video only one known tracker is expected to perform the best.
Using the labeled video, we then fine-tune a classifier \cite{dosovitskiy2020image, caron2021emerging} on top of the powerful CNN-based and ViT-based backbone architectures including MoCo V2 \cite{chen2020improved}, Barlow Twins \cite{zbontar2021barlow}, SwAV \cite{caron2020unsupervised}, and DINO \cite{caron2021emerging} in an end-to-end fashion, which we then use for the selection of a particular tracker at test time.
TP$^{2}$N is trained on a compound dataset obtained from the training splits of LaSoT, GOT-10K, and TrackingNet.
Therefore, it generalizes well to a wide range of tracking scenarios.
Each video sequence is labeled by using 17 SOTA trackers making our TP$^{2}$N quite comprehensive across varying tracking paradigms.
At inference time, these 17 SOTA trackers are not required to be executed while our TP$^{2}$N will predict the best-performing tracker.
It significantly improves over the existing practice of executing multiple trackers at inference time and selecting the best performer \cite{ZHAO2021107679, dunnhofer2022combining, tang2023exploring, LEANG2018459}.
We observe that executing multiple trackers is not necessary, as the proposed TP$^{2}$N is applicable in a wide range of conditions, both at the sequence and the frame levels. 

The proposed method is based on \textit{selecting a tracker} which is predicted to perform well as opposed to the more general approach of tracker output {\it combination}. First, tracker combination requires running multiple trackers and thus slows the meta-tracker. Second, many parameters would have to be pre-trained or estimated at test time, e.g. the weights of individual trackers and the method of output combination. The two disadvantages, together with the good experimental performance lead to our decision to focus on tracker selection.%
\footnote{We acknowledge that in the per-frame selection mode, the BofN meta-tracker runs all $N$ trackers. The optional per-frame mode is described in Sec. \ref{test}.}

We show in the experimental section that if tracking output is selected on a per-frame basis, further non-negligible improvement is achieved.
We conjecture that because different VOT paradigms are based on different assumptions, and as stated in the `No Free Lunch Theorem' \cite{585893}, no single visual tracker works best in every tracking situation.

The proposed TP$^{2}$N is evaluated on the test splits of the publicly available VOT benchmark datasets including LaSOT \cite{8954084}, TrackingNet \cite{Muller_2018_ECCV}, GOT10K \cite{8922619}, VOT2021 \cite{Kristan_2021_ICCV} and VOT2022 \cite{10.1007/978-3-319-10599-4_13}.
The potentially best tracker may be predicted using TP$^{2}$N both at the video-level and the frame-level.
In the case of long-term tracking, the target and scene characteristics may significantly vary within the same video, requiring trackers to switch within a sequence requiring frame-level prediction.
To handle this issue, we experimented by applying the TP$^{2}$N, multiple times within the same video to ensure the initially selected tracker was appropriate at different time steps.
 Our results demonstrate a significant performance improvement over all datasets compared to the baseline SOTA trackers, thus verifying our proposed selection strategy for the best of the \textit{N} trackers.
We summarize our main contributions as follows:
\begin{enumerate}
    \item We propose a Tracker Performance Prediction Network (TP$^{2}$N) which may be employed at
    varying temporal resolutions, including video-level and frame-level.
    \item The proposed ``Best of the $N$ trackers'' has minimal overhead compared to other tracker ensembling paradigms \cite{ZHAO2021107679, dunnhofer2022combining, tang2023exploring, LEANG2018459}.
    \item Large-scale experiments are performed on several publicly available benchmark datasets and demonstrate the superior performance of the proposed best of the $N=17$ trackers. 
\end{enumerate}

The rest of the paper is structured as follows: Sec. \ref{sec:relatedwork} provides a detailed review of related works on different tracer fusion approaches. 
Sec. \ref{sec:proposed} explains the proposed algorithm in detail.
Sec. \ref{sec:results} presents an exhaustive evaluation of the proposed algorithm, while Sec. \ref{sec:conclusion} delivers the conclusion and outlines potential trajectories for future exploration in this domain.

\section{Related Work}
\label{sec:relatedwork}


To handle wide variations in tracking performance, different types of fusion approaches have been developed in the literature, including early and late fusion, decision-level fusion, and ensembling of SOTA trackers \cite{tang2023exploring}.

\noindent \textbf{Early Fusion:} In early fusion, different types of feature representations capturing complementary information about the target object are concatenated together and input into the tracking paradigm for improved tracking performance \cite{ZHAO2021107679, Li_2019_CVPR, 9018389, Xu_2019_ICCV, Qi_2016_CVPR, Bhat_2018_ECCV, Fan_2019_CVPR, He_2018_CVPR, nam2016mdnet, chen2019multi}.
The feature-level fusion realizes the fusion process in an embedding space, obtained by an integrated feature extraction performed by a deep neural network.
For instance, the MDLatLRR tracker fused RGB  and IR images using a nuclear norm-based constraint \cite{9018389}.
The UPDT tracker learns the weights of the individual feature representations for feature fusion \cite{Bhat_2018_ECCV}.
Xu \textit{et al.} proposed a GFS-DCF tracker for group feature selection in correlation filters using low-rank regularization \cite{Xu_2019_ICCV}.
More features-level fusion approaches for improving VOT include SiamRPN++ \cite{Li_2019_CVPR}, C-RPN \cite{Fan_2019_CVPR}, SA-Siam \cite{He_2018_CVPR}, MDNET \cite{nam2016mdnet}, and MAM  \cite{chen2019multi}.
In early fusion, different types of features are fused to improve the VOT performance, however, a particular feature type may not be effective for a particular tracking scenario resulting in less than optimal performance.
For each tracking scenario, detecting the optimal set of effective feature types is a computationally intractable problem.

\noindent \textbf{Late Fusion:} In late fusion or decision-level fusion, the tracker responses are estimated by varying feature representations as well as different tracker types.
The responses obtained by different trackers are then fused for improved VOT performance \cite{10.1007/978-3-319-10599-4_13, Ma_2015_ICCV}.
For instance, the MEEM tracker uses historical information and constitutes an expert ensemble, where the best expert is selected to restore the current tracker based on a minimum entropy criterion \cite{10.1007/978-3-319-10599-4_13}.
HCF tracker employs hierarchical deep feature representation and trains correlation filters on each layer \cite{Ma_2015_ICCV}.
Different correlation filters are then fused using a weighting factor to perform VOT.
The MCCT tracker also employs the approach where multiple proposals based on the agreement of multiple feature combinations as well as temporal consistency are selected \cite{8578607}. 
Compared to early fusion, late fusion requires multiple responses to be estimated in parallel on a single video sequence, resulting in increased computational complexity \cite{tang2023exploring}.


\noindent \textbf{Ensembling of SOTA Trackers:} In late fusion, the set of features or trackers is fixed beforehand, while a badly performing feature or tracker may result in reduced overall performance.
To handle this issue, tracker ensembling has also been proposed to select the best-performing trackers for late fusion.
For this purpose, different types of approaches have been proposed to estimate the best performer in a given tracking scenario \cite{Choi_2017_CVPR, 8667882, Dai_2020_CVPR, LEANG2018459}.
In the ACFN tracker, an attention mechanism is used to predict the best-performing tracker for later fusion \cite{Choi_2017_CVPR}.
PTAV integrates SINT \cite{Tao_2016_CVPR} and fDSST \cite{7569092} into an efficient and accurate tracker using a tracking-validation framework \cite{8667882}. 
Dunnhofer \textit{et al.} uses confidence scores to select the best-performing trackers for late fusion \cite{dunnhofer2022combining}.
Similarly, Dai \textit{et al.} employed a meta-updater to predict the accuracy of the different trackers in the ensemble while the low-performing trackers are removed \cite{Dai_2020_CVPR}.
Leang \textit{et al.} employed a drift prediction mechanism for estimating the accuracy of the trackers to be fused \cite{LEANG2018459}.

Although the best trackers are fused, all trackers have to be executed for inference, resulting in increased computational complexity and limiting the potential set of trackers to be employed \cite{tang2023exploring}.
Most of these fusion approaches have no mechanism to predict the performance of a particular tracker in a given scenario without actually executing that tracker.
In contrast, we propose a mechanism to predict the potential performance of a given tracker in a particular tracking scenario without actually executing that tracker.
To the best of our knowledge, we are the first ones to propose a mechanism for predicting the best tracker for a specific tracking scenario.
Our proposed ``Best of the $N$ trackers'' is simple and efficient for quick scene-level tracker selection.
The overhead is minimal and the performance remains the same as that of the best-predicted tracker.

\begin{figure*}[t!]
\centering
\includegraphics[width=\linewidth]{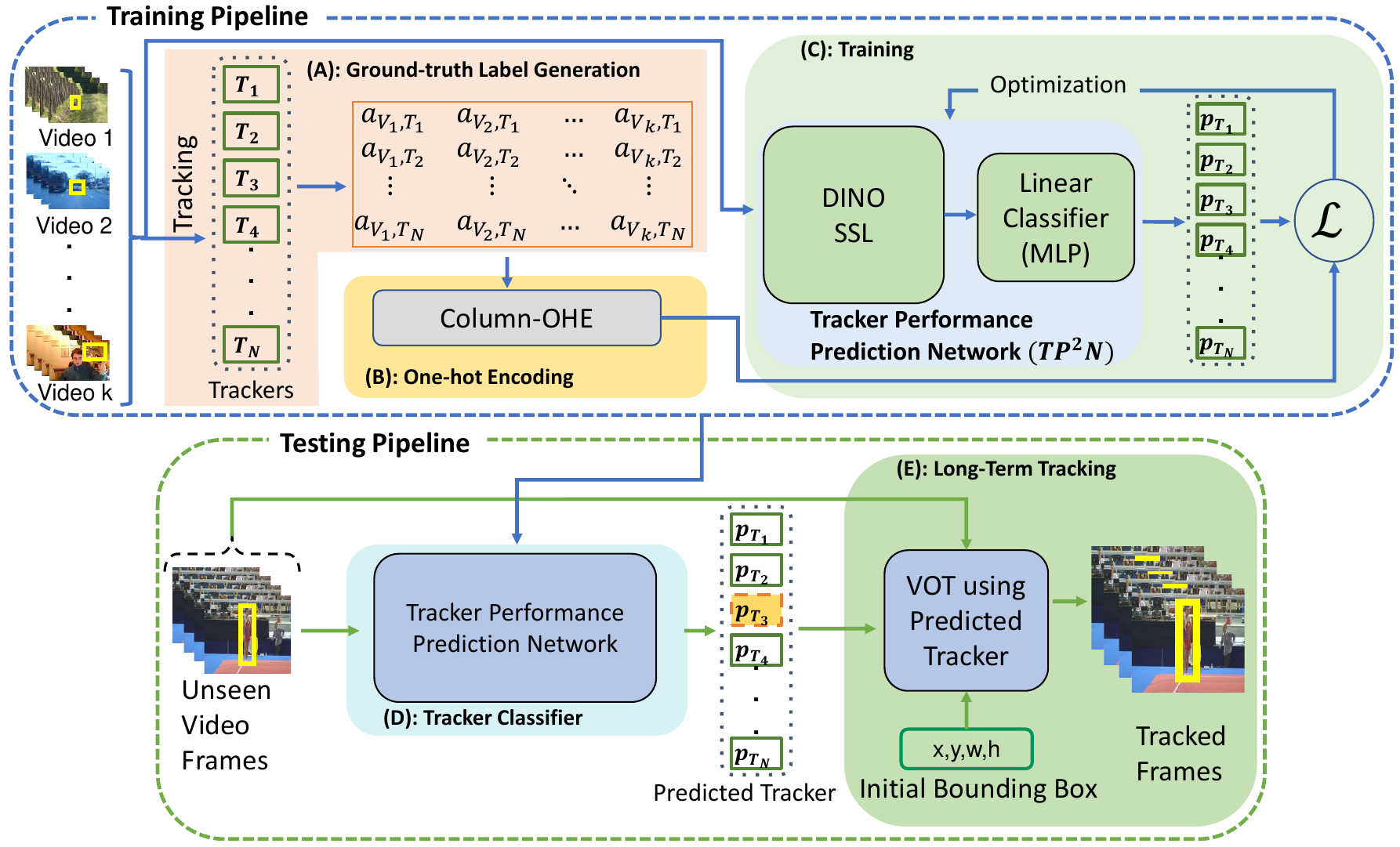}
\label{fig:first}
\caption{Structure of the proposed BofN -- ``Best of the $N$ Trackers''. }
\label{fig2}
\end{figure*}

\section{Proposed Methodology}
\label{sec:proposed}
The system diagram of the proposed ``Best of the $N$ Trackers'' (BofN) meta-tracker is shown in Fig. \ref{fig2}.
It consists of two major steps, including groundtruth label generation from the input video sequence using a set of SOTA visual trackers (Sec. \ref{groundtruth} $\&$ Fig. \ref{fig2} (steps (A)-(B))) and training a Tracking Performance Prediction Network (T$P^{2}$N) (Sec. \ref{training} $\&$ Fig. \ref{fig2} (step (C))).
\vspace{-0.5cm}
\subsection{Problem Formulation}
Given a set of $K$ training video sequences, $\mathcal{V}=\{V_{1}, V_{2},\cdot \cdot \cdot, V_{K} \}$, where $V_{j}=\{f_{jk}\}_{k=1}^{m_j}$, $m_j$ is the number of frames in each video $V_j$.  
Each frame $f_{jk}$ in $V_j$ is annotated with a bounding box/segmentation mask: $\mathcal{B}_j=\{b_{j1}, b_{j2},\cdot \cdot \cdot, b_{jm} \}$.
In the case of a bounding box, $b_{jk}=[x, y, w, h]$ represents the top left corner coordinates with the width and height of the target object.
In the case of segmentation-based annotation, $b_{jk}$ represents a segmentation mask of the target object.

Given a set of $N$ visual trackers, $\mathcal{T}=\{T_{1}, T_{2},\cdot \cdot \cdot, T_{N} \}$, our task here is to predict the best-performing tracker $T_{i}$ for a test video sequence $V_{j}$ such that only the bounding box $b_{j1}$ in the first frame is available at the inference stage.
For this purpose, we propose to train a tracking performance prediction network $f(\theta)$ that will take an input scene from a video sequence and predict the best-performing tracker: $f(\theta): S_j \longrightarrow T_i$, where $S_j$ is the subsequence consisting of two or more frames of the video $V_j$ and a bounding box $b_{ij}$.

\subsection{Groundtruth Label Generation (Figs. \ref{fig2} (A)-(B))}
\label{groundtruth}
To train the TP$^{2}$N as $f(\theta)$, each of the trackers $T_i \in \mathcal{T}$ is executed on each training video $V_j \in \mathcal{V}$, and its performance in terms of success rate (AUC) is computed.
The performance of all trackers is arranged in a vector $\textbf{a}_{j}=[a_{V_{j}, T_{1}},a_{V_{j}, T_{2}},\cdots, a_{V_{j}, T_{N}}]^{\top}$, where $a_{V_{j}, T_{i}}$ is the performance of a  tracker $T_i$ on a video $V_j$.
We change this vector to probabilities by dividing it to its magnitude $\textbf{p}_{j}=\textbf{a}_{j}/||\textbf{a}_{j}||_{2}$.
For the classification problem, the vector $\textbf{p}_{j}$ is thresholded by mapping maximum probability to 1 and others to 0, resulting in a One Hot Encoding (OHE) representation.
These groundtruth labels are used during the training of our proposed TP$^{2}$N.

\subsection{Fine-tuning the Proposed Tracking Performance Prediction Network (TP$^{2}$N) (Fig. \ref{fig2} (C))}
\label{training}
The proposed TP$^{2}$N may utilize any existing architecture as a classifier or regressor.
However, to cope with limited available datasets as developed by groundtruth label generator step, we employ Self-Supervised Learning (SSL) \cite{chen2020simple} based paradigms such as CNNs-based architectures including MocoV2 \cite{chen2020improved}, SwaV \cite{caron2020unsupervised}, and ViT-based architecture known as DINO \cite{caron2021emerging}. 
We also employ BarlowTwins \cite{zbontar2021barlow} as a non-contrastive learning method to pre-train our TP$^{2}$N.
These architectures are pre-trained using a large-scale ImageNet dataset in SSL settings \cite{chen2020simple}.
These methods learn robust representations using pre-text tasks exploiting supervisory signals obtained from the unlabeled data. 

We further fine-tune these architectures using the proposed dataset for the ``Best of the $N$ Trackers'' prediction.
Using the well-established SSL evaluation protocols, the fine-tuning is performed in two different settings including linear probing and full fine-tuning.
In linear probing, the backbone feature extractor is kept frozen while the remaining parts of the model (e.g., linear
classifier or decoders) is trained for the best tracker prediction task.
In full fine-tuning (denoted as Fine-tune) evaluation, all layers including the backbone are fine-tuned.
The linear probing depends more strongly on the quality of the learned representations whereas in full fine-tuning the performance is more affected by the transferability of the learned weights.

\subsection {Data Augmentation}
The performance of SSL architectures may vary depending on the composition of training data and the selected set of data augmentation methods. 
In order to adapt the SSL architectures for the task of the ``Best of the $N$ trackers'' prediction, we employ several VOT-specific data augmentation techniques including temporal resolution reduction, spatial resolution reduction, backward video sequence adaptation, and target object scale variations.
In addition to that, we also employ often-used image processing augmentation techniques including random orientations, random flipping, and color distortion.

In temporal resolution reduction, temporal sub-sampling is performed by varying the frame rates randomly as 10\% and 50\%.
In spatial resolution reduction, spatial sub-sampling is performed by a factor of 10\% and 50\%.
For backward video adaptation, visual trackers are applied in opposite temporal directions.
For target scale variations, the target bounding boxes are randomly scaled down using scale factors of 10\%, 20\%, and 50\%.
As a result of these augmentations, we introduce significant diversity in our training data resulting in 10 fold increase in the number of training samples.

\subsection {Inference using TP$^{2}$N (Fig. \ref{fig2} (D)-(E))}
\label{test}
We apply TP$^{2}$N on unseen test video sequences in two different settings including video-level and frame-level.
In video-level prediction, we predict the ``Best of the $N$ Trackers'' only once for each video using the initial frames of that video.
In the frame-level prediction, the TP$^{2}$N is applied after fixed temporal intervals such that the prediction overhead does not become dominant while detecting a change in the scene characteristics or the object being tracked as quickly as possible.
Such variations may require a switching ``Best of the $N$ Trackers'' resulting in performance improvement.
In the experimental results, we observe that frame-level prediction is more accurate than video-level prediction, especially in the long-term VOT sequence.
However, it will also incur more overhead due to the increased prediction frequency.

\begin{table*}[t!]
    \centering
    \caption{\textbf{
    Tracking performance prediction network (TP$^2$N) -- evaluation at the video level} 
    in terms of Top-1 accuracy for both linear and fine-tuning experiments on the test split of each dataset. The models are trained using the union of the training split of the dataset. Note that $p$ represents the patch size used in ViT architecture.}
    \begin{tabular}{ c c| c c| c c| c c| c c }
    \hline
    \textbf{Arch.} & \textbf{Method} & \multicolumn{2}{|c|}{\textbf{LaSOT}} & \multicolumn{2}{c|}{\textbf{TrackingNet}} & \multicolumn{2}{c|}{\textbf{GOT-10k}} & \multicolumn{2}{c}{\textbf{VOT2022}}\\
    & & Linear & Fine-tune & Linear & Fine-tune & Linear & Fine-tune & Linear & Fine-tune \\
    \hline
    \multirow{3}{*}{ResNet-50} & MoCo v2 &0.808 &\underline{0.911}&0.833 &0.902 &0.865 &0.912&\textbf{0.843}&0.908\\
     & SwAV &0.821&0.851 &0.851&0.881&0.881&0.924& 0.791& 0.867\\
     & BT &\textbf{0.861}&0.879&0.842&0.865&0.844&0.892&0.829&0.882\\
    \hline
    \multirow{2}{*}{ViT-S} & DINO\textsubscript{p=16}&\underline{0.851}&0.897&\underline{0.871}& \underline{0.937}&\textbf{0.902}&\textbf{0.962}&\underline{0.841}&\underline{0.923}\\
     & DINO\textsubscript{p=8} &0.842&\textbf{0.945}&\textbf{0.897}&\textbf{0.966}&\underline{0.893}&\underline{0.934}&0.809&\textbf{0.955} \\
    \hline
    \end{tabular}
    \label{table1}
\end{table*}

\begin{table*}[t!]
    \centering
   \caption{\textbf{
    Tracking performance prediction network (TP$^2$N)-- evaluation at the frame level} 
    in terms of Top-1 accuracy for both linear and fine-tuning experiments on the test split of each dataset. The models are trained using the union of training split of the evaluated dataset. Note that $p$ represents the patch size used in ViT architecture.}
    \begin{tabular}{ c c| c c| c c| c c| c c }
    \hline
    \textbf{Arch.} & \textbf{Method} & \multicolumn{2}{|c|}{\textbf{LaSOT}} & \multicolumn{2}{c|}{\textbf{TrackingNet}} & \multicolumn{2}{c|}{\textbf{GOT-10k}} & \multicolumn{2}{c}{\textbf{VOT2022}}\\
    & & Linear & Fine-tune & Linear & Fine-tune & Linear & Fine-tune & Linear & Fine-tune \\
    \hline
    \multirow{3}{*}{ResNet-50}&MoCo v2&\underline{0.750}&\underline{0.813}&0.706&0.792&0.811&0.843&0.792&0.831\\
     & SwAV &0.746&0.771&0.671&0.724&0.786&0.801&0.866&0.887\\
     & BT &\textbf{0.771}&0.793&0.701&0.753&0.832&0.859&0.845&0.883\\
    \hline
    \multirow{2}{*}{ViT-S} & DINO\textsubscript{p=16}&0.743&0.802&\underline{0.788}&\underline{0.863}&\underline{0.855}&\underline{0.901}&\textbf{0.903}&\textbf{0.953}\\
     & DINO\textsubscript{p=8}&0.721&\textbf{0.833}&\textbf{0.821}&\textbf{0.892}&\textbf{0.879}&\textbf{0.936}&\underline{0.887}&\underline{0.945}\\
    \hline
    \end{tabular}
    \label{table2}
\end{table*}

\begin{table*}[t!]
\centering
\caption{BofN meta-tracker -- architecture selection.
Fine-tuned performance, both at the Frame and Video levels, on  LaSOT and TrackingNet (AUC reported), GOT-10K (AO, average overlap), and on VOT (EAO, expected average overlap). All the metrics are the primary ones for the given dataset.}
\begin{tabular}{ c c| c c| c c| c c| c c }
\hline
\textbf{BofN} & \textbf{Method} & \multicolumn{2}{|c|}{\textbf{LaSOT}} & \multicolumn{2}{c|}{\textbf{TrackingNet}} & \multicolumn{2}{c|}{\textbf{GOT-10k}} & \multicolumn{2}{c}{\textbf{VOT2022}}\\
\textbf{Tracker}& & Frame & Video & Frame & Video & Frame & Video & Frame & Video \\
\hline
\multirow{5}{*}{ResNet-50} & MoCo v2 &\textbf{80.91}&76.80&\textbf{91.17}&\textbf{87.78}&\underline{89.76}&\underline{87.10}&69.46&\textbf{65.85}\\
& SwAV &79.11&74.50&90.50&\underline{87.50}&89.51&\textbf{87.98}&\textbf{70.00}&65.32\\
& BT &79.76&75.10&90.89&87.10&\textbf{90.14}&86.40&\underline{69.76}&\underline{65.71}\\
& DINO\textsubscript{p=16}&79.19&\underline{77.96}&90.77&86.77&87.60 &83.20 &66.70 &63.47 \\
    & DINO\textsubscript{p=8}&\underline{80.62}&\textbf{78.75}&\underline{91.10}&84.64&85.70&83.30 &66.30&64.98 \\
\hline
\multirow{5}{*}{ViT-S} & MoCo v2 &78.81&72.22&88.61&85.97&84.10&81.31&64.55&61.74 \\
     &SwAV &\underline{80.64}&\underline{75.64}&91.58&85.10&83.64&80.78&62.33&59.54 \\
     &BT&76.60&72.48&89.65&86.54&85.61 &82.47&63.44 &60.11 \\
     & DINO\textsubscript{p=16} & 80.65&75.60&\underline{92.03} &\underline{88.70}&\underline{91.08} &\textbf{88.76}&\textbf{70.98}&\underline{67.52}\\
& DINO\textsubscript{p=8} &\textbf{81.77}&\textbf{77.60}&\textbf{92.10}&\textbf{88.90}&\textbf{91.12}&\underline{88.50}&\underline{70.17}&\textbf{67.88}\\
\hline
\end{tabular}
\label{table3}
\end{table*}

\section{Experimental Evaluations}
\label{sec:results}
In order to evaluate the performance of the BofN meta-tracker, a large number of experiments are performed on four different datasets including LaSOT \cite{8954084}, GOT-10K \cite{8922619}, TrackingNet \cite{Muller_2018_ECCV}, and VOT2022 \cite{Kristan_2022_ICCV}.
We employed 17 SOTA best-performing trackers for the purpose of data annotation including RTS \cite{Paul2022}, TOMP \cite{mayer2022transforming}, AiATrack \cite{gao2022aiatrack}, DropTrack \cite{dropmae2023}, OSTrack \cite{ye2022ostrack}, GRM \cite{gao2023generalized}, CiteTracker \cite{li2023citetracker}, ARTrack \cite{Wei_2023_CVPR},  SiamFC \cite{bertinetto2016fully}, SiamRPN++ \cite{Li2019a}, Ocean \cite{Ocean_2020_ECCV}, TransT \cite{Chen2021}, KeepTrack \cite{Mayer2021b}, STARK \cite{Yan2022}, SimTrack \cite{chen2022backbone}, MixFormer \cite{cui2022mixformer}, and SwinTrack \cite{lin2021swintrack}.

\subsection{Implementation Details}
We fine-tuned our Tracking Performance Prediction Network (TP$^{2}$N) by combining the training splits of  TrackingNet (30130 sequences), LaSOT (1120 sequences), and GOT-10K (9340 sequences) datasets.
We employ four different SSL backbones including ResNet-50 architecture for MoCo V2 \cite{chen2020improved}, Barlow Twins (BT) \cite{zbontar2021barlow}, and SwAV \cite{caron2020unsupervised}.
For DINO, we employ ViT-small (ViT-S) with $8 \times 8$ and $16 \times 16$ patch sizes.
For each SSL method for fine-tuning the TP$^{2}$N, the same protocols are used as proposed by the original authors.
The scaling rule is used to adjust the learning rate: $l_{r}=l_{r}*batchsize/256$.
Each backbone is pre-trained on 4 V100 Tesla GPUs for 200 ImageNet epochs as a pre-training step \cite{tian2021divide}.
We follow the same data pre-processing and training schemes as in \cite{graham2019simultaneous}.

For MoCoV2 \cite{chen2020improved}, SGD optimizer is used with an initial learning of 0.3 which is linearly scaled using the aforementioned learning scheme in which the batch size is 4,096.
The memory bank size is fixed to 65,536, and momentum of 0.999 is used with a weight decay of $10^{-4}$ is used for regularization.
For SwAV \cite{caron2020unsupervised}, the optimizer and learning rate are the same while the batch size is 2,048.
For BT, LARS optimizer is used with a learning rate of 0.2 for weights and 0.0048 for biases as recommended by the original authors \cite{you2017large}.
The embedding dimension is 8, 192 and training is performed wiht a coefficient of off-diagonal term $\lambda=5. 10^{-3}$ with a weight decay of $\lambda=1. 5. 10^{-6}$
For both variants of the DINO ($8 \times 8$ and $16 \times 16$ patch sizes), AdamW optimizer is used with a learning rate of 0.0005 which is linearly scaled $l_{r}=l_{r}*batchsize/256$ with a batch size of 1,024.
Weight decay follows a cosine schedule form 0.04 to 0.4.

\subsection{Datasets}
The following VOT datasets are used in this work:

\begin{itemize}
    \item \textbf{LaSOT} \cite{8954084}: LaSOT is a large-scale benchmark dataset that is mostly employed for long-term object tracking performance. 
    The official training split contains 1120 sequences (2.8M frames) and the testing split contains 28 sequences (685K frames). 
    \item \textbf{TrackingNet} \cite{Muller_2018_ECCV}: The TrackingNet contains a wide variety of real-world object types and tracking challenges. 
    The dataset contains a training split of 30,130 (14M frames) sequences and testing split contains 511 (226k frames) sequences. 

    \item \textbf{GOT-10k} \cite{8922619}: The GOT-10K dataset contains a training split of 9340 (1.4M frames) sequences while the testing split has 420 sequences (56K frames). This dataset contains a wide variety of tracking challenges and no overlapped sequence between the training and testing splits.
     \item \textbf{VOT2019 \textcolor{black}{\cite{VOT2019}}, VOT2021 \cite{Kristan_2021_ICCV}, \& VOT2022 \cite{Kristan_2022_ICCV}}: The VOT datasets series consists of 60 video sequences, where the frames are in RGB format. 
     The VOT2021 and VOT-2022 contain 19.447k and 19.903k frames, respectively. We employ these datasets for evaluation purposes.
      \item \textbf{OTB100} \textcolor{black}{\cite{otb100}}: The OTB100 dataset contains 100 challenging sequences with 58,897 frames and 356 $\times$ 530 average resolution. This  dataset also contains 11 distinct attributes.
      \item \textbf{UAV123} \textcolor{black}{\cite{mueller2016benchmark}}: This dataset contains 123 UAV sequences with diverse tracking attributes. There are total 21.9K frames with an average resolution of 1280 $\times$ 720.
\item \textbf{WebUAV-3M} \textcolor{black}{\cite{webuav_3M}}: The WebUAV-3M dataset contains 4500 video sequences and offers 223 highly diverse target categories with over 3.3 million frames.
The training/validation/testing splits contain 3520/200/780 videos.
We only evaluate our proposed tracker using the testing split which shows its generalization capability. 
   
\end{itemize}

\subsection{Evaluation Metrics}
Similar to the existing SOTA VOT methods \cite{Wei_2023_CVPR, Kristan_2022_ICCV}, we also employed evaluation metrics including Area Under Curve (AUC), Precision (P), Normalized Precision ($P_{Norm}$), Average Overlap (AO), Success Rate ($SR_{0.75}$), $SR_{0.5}$, Expected Average Overlap (EAO), Robustness (R), and Accuracy (A) for performance comparison. 
The performances of VOT2021-2022 datasets are reported using the EAO, A, and R measures while the rest of other metrics are employed for other datasets.
Moreover, for evaluating the performance of TP$^{2}$N, we used top-1 classification accuracy for both frame-level and video-level experiments.

\subsection{Evaluation of TP$^{2}$N}
We perform various experiments for the evaluation of TP$^{2}$N for prediction of the best-performing tracker.
We compare two different architectures ResNet-50 and ViT-S for representation learning and four different SSL methods including MoCo V2, SWAv, BT, and DINO in two settings ($p=8$ and $p=16$).
Two different types of fine-tunings are employed including Full-fine tuning (Fine-Tune) and Linear probing (Linear) using three linear layers.
All methods are evaluated on the testing splits of the four VOT benchmarks including LaSOT, TrackingNet, GOT-1OK, and VOT2022.

Table \ref{table1} shows the video-level performance evaluation of TP$^{2}$N.
Overall, the full fine-tuning experiments have obtained more performance than the corresponding linear probing results.
Specifically, on LaSOT dataset, DINO$_{p=8}$ with full fine-tuning has obtained a maximum top-1 classification performance of 94.50$\%$ while MoCo V2 has obtained second best performance of 91.10$\%$.
On TrackingNet, DINO$_{p=8}$ has obtained the best performance of 96.60$\%$ while DINO$_{p=16}$ obtained the second best performance of 93.70$\%$.
On GOT-1OK, DINO$_{p=16}$ has obtained the best performance of 96.20$\%$ while DINO$_{p=8}$ has remained the second best performer with 93.40$\%$.
Similarly, on VOT2022 dataset, DINO$_{p=8}$ has obtained the best performance of 95.50$\%$ while DINO$_{p=16}$ remained the second best performer with 92.30$\%$.

Table \ref{table2} shows the frame-level performance of the TP$^{2}$N with different SSL methods and different backbone architectures.
It is observed that a full fine-tuning evaluation protocol has obtained a better performance compared to linear probing in all four datasets.
Also, the DINO SSL method has remained the first and second-best performer on all datasets.

\begin{figure*}[t!]
\centering
\includegraphics[width=\linewidth]{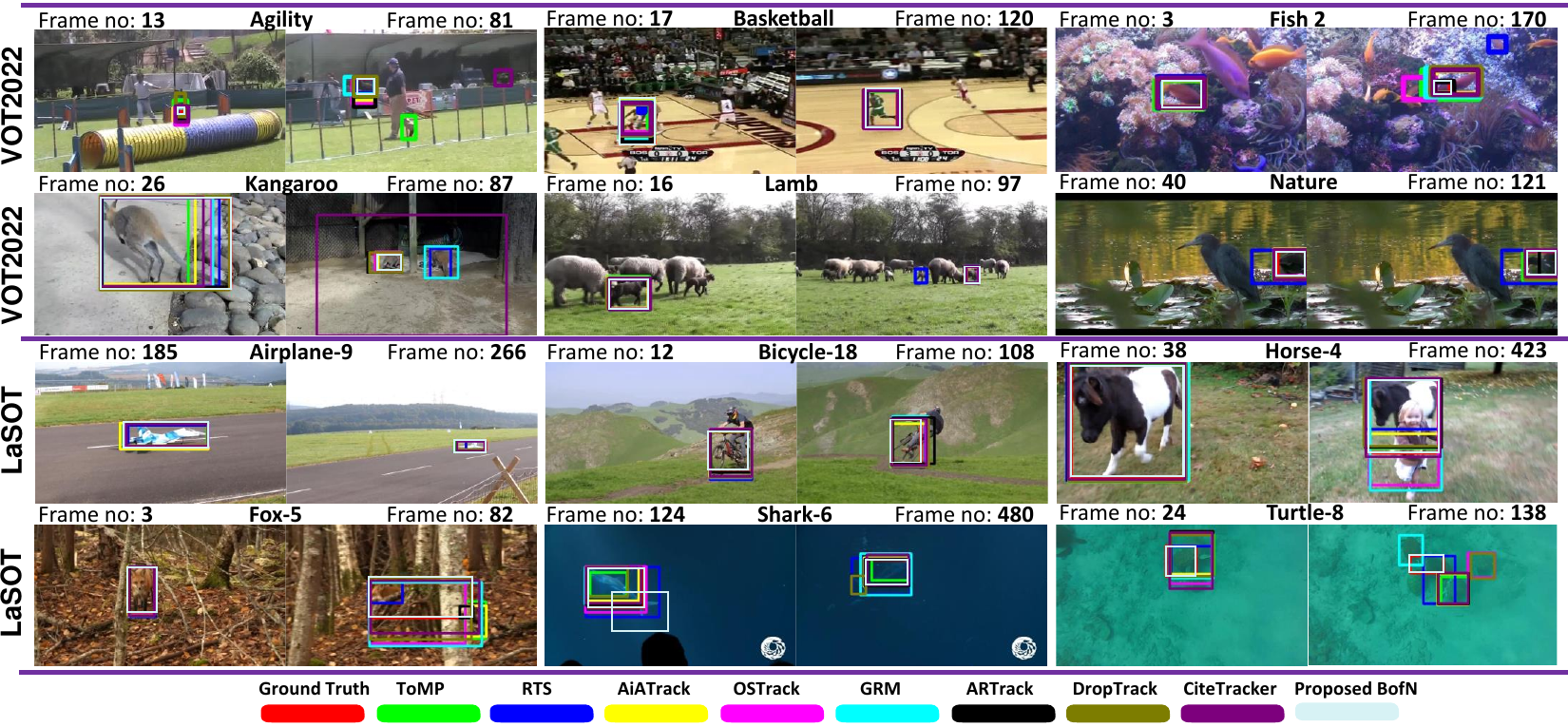}
\label{fig:first}
\caption{Visual results of the proposed BofN -- ``Best of the $N$ Trackers'' and its comparison with existing SOTA trackers, including ToMP \cite{mayer2022transforming}, RTS \cite{Paul2022}, AiATrack \cite{gao2022aiatrack}, OSTrack \cite{ye2022ostrack}, GRM \cite{gao2023generalized}, ARTrack \cite{Wei_2023_CVPR}, DropTrack \cite{dropmae2023}, and CiteTracker \cite{li2023citetracker} on 12 challenging sequences selected from LaSOT \cite{8954084} and VOT2022 \cite{Kristan_2022_ICCV} 
datasets.
Frame indexes and sequence names are shown for each sequence. 
Our proposed BofN tracker has consistently performed well against these challenges as compared to the other trackers.}
\label{fig_visual}
\end{figure*}

\begin{table*}[t!]
\centering
\caption{\textbf{BofN meta-tracker, with ViT DINO\textsubscript{p=8} model in TP$^2$N}
-- comparison with state-of-the-art trackers on  LaSOT \cite{8954084}, TrackingNet \cite{Muller_2018_ECCV}, and GOT-10k \cite{8922619} datasets. 
The best results are in bold and the second best results are underlined.
{For model selection in TP$^2$N, the tracker prediction network, see Table \ref{table3}.} 
}
\begin{tabular}{ c | c c c | c c c | c c c}
\hline
\multirow{2}{*}{Method} & \multicolumn{3}{c|}{LaSOT \cite{8954084}} & \multicolumn{3}{c|}{TrackingNet \cite{Muller_2018_ECCV}} & \multicolumn{3}{c}{GOT-10k \cite{8922619}}  \\
&AUC& $P_{Norm}$ & P &AUC& $P_{Norm}$&P&AO& $SR_{0.75}$ & $SR_{0.5}$  \\
\hline
SiamFC \hfill (2016)\cite{bertinetto2016fully} & 33.6 & 42.0 & 33.9 & 57.1 & 66.3 & 53.3 & 34.8 & 9.8 & 35.3 \\
SiamRPN++ \hfill  (2019) \cite{Li2019a} & 49.6 & 56.9 & 49.1 & 73.3 & 80.0 & 69.4 & 51.7 & 32.5 & 61.6 \\
Ocean \hfill (2020) \cite{Ocean_2020_ECCV} & 56.0 & 65.1 & 56.6 & 71.2 & 78.6 & 65.9 & 61.1 & 47.3 & 72.1 \\
TransT \hfill (2021) \cite{Chen2021} & 64.9 & 73.8 & 69.0 & 81.4 & 86.7 & 80.3 & 67.1 & 60.9 & 76.8 \\
KeepTrack \hfill (2021) \cite{Mayer2021b} & 67.1 & 77.2 & 70.2 & 80.0 & 84.1 & 79.0 & 63.0 & 58.2 & 70.8 \\
STARK \hfill (2021)\cite{Yan2022} & 67.1 & 77.0 & 73.2 & 82.0 & 86.9 & 76.8 & 68.8 & 64.1 & 78.1 \\
SimTrack \hfill (2022) \cite{chen2022backbone} & 70.5 & 79.7 & 74.9 & 83.4 & 87.4 & 81.8 & 69.0& 65.4 & 71.8\\
MixFormer \hfill (2022) \cite{cui2022mixformer} & 70.1 & 79.9 & 76.3 & 83.9 & 88.9 & 83.1 & 71.2 & 65.8 & 79.9 \\
SwinTrack \hfill (2022) \cite{lin2021swintrack} & 71.3 & 80.7 & 76.5 & 84.0 & 86.8 & 82.8 & 72.4 & 67.8 & 80.5 \\
RTS \hfill (2022) \cite{Paul2022} & 69.7 & 76.2 & 73.7 & 81.6 & 86.0 & 79.4 & 85.2 & 82.6 & 94.5 \\
TOMP \hfill (2022)  \cite{mayer2022transforming} & 68.5 & 79.2 & 73.5 & 81.5 & 86.4 & 78.9 & 70.8 & 66.2 & 76.8 \\
OSTrack \hfill  (2022) \cite{ye2022ostrack} & 71.1 & 81.1 & 77.6 & 83.9 & 88.5 & 83.2 & 73.7 & 70.8 & 83.2 \\
AiATrack \hfill (2022) \cite{gao2022aiatrack} & 69.0 & 79.4 & 73.8 & 82.7 & 87.8 & 80.4 & 69.6 & 63.2 & 80.0 \\
DropTrack \hfill (2023) \cite{dropmae2023} & 71.8 & 81.8 & 78.1 & 84.1 & 88.9 & 85.4& 75.9 & 72.0 & 86.8 \\
GRM \hfill (2023) \cite{gao2023generalized} & 69.9 & 79.3 & 75.8 & 84.0 & 88.7 & 83.3 & 73.4 & 70.4 & 82.9 \\
CiteTracker \hfill (2023) \cite{li2023citetracker} & 69.7 & 78.6 & 75.7 & 84.5 & 89.0 & 84.2 & 74.7 & 73.0 & 84.3 \\
ARTrack \hfill (2023) \cite{Wei_2023_CVPR} & 73.1 & 82.2 & 80.3 & 85.6 & 89.6 & 86.0 & 78.5 & 77.8 & 87.4 \\
\hline
\multirow{3}{*}{}\textbf{BofN meta-Tracker} & \multicolumn{3}{c|}{LaSOT \cite{8954084}} & \multicolumn{3}{c|}{TrackingNet \cite{Muller_2018_ECCV}} & \multicolumn{3}{c}{GOT-10k \cite{8922619}}  \\
\cline{2-10}
Video-Level&\underline{77.6}&\underline{85.8}&\underline{83.7}&\underline{88.9}&\underline{93.4}&\underline{89.7}&\underline{88.7}&\underline{85.20}&\underline{95.31}  \\
Frame-Level&\textbf{81.7}&\textbf{87.6}&\textbf{86.2}&\textbf{92.1}&\textbf{96.6}&\textbf{92.5}&\textbf{91.1}&\textbf{88.56}&\textbf{97.02}\\
\hline
\end{tabular}
\label{table4}
\end{table*}

\begin{table*}[t!]
    \centering
   \caption{\textbf{BofN meta-tracker} -- comparison with state-of-the-art trackers on VOT2021-2022 \cite{Kristan_2021_ICCV, Kristan_2022_ICCV} datasets. 
The best results are in bold and the second best results are underlined.
{For model selection in TP$^2$N, the tracker prediction network, see Table \ref{table3}.} 
}
    \begin{tabular}{ l | c c c | c c c }
    \hline
    \multirow{2}{*}{Method} & \multicolumn{3}{c|}{VOT2021 \cite{Kristan_2021_ICCV}} & \multicolumn{3}{c}{VOT2022 \cite{Kristan_2022_ICCV}} \\
    &EAO&A&R&EAO&A&R  \\
    \hline
    SiamFC \hfill (2016)\cite{bertinetto2016fully}&26.61 & 55.10& 52.90& 25.50 & 56.20 & 54.30 \\
    SiamRPN++  \hfill (2019)\cite{Li2019a} & 48.88& 64.50& 68.20&53.20 &69.20 &70.10  \\
    Ocean  \hfill (2020) \cite{Ocean_2020_ECCV} &50.88& 68.20&72.50& 59.23&73.20 & 74.20\\
    TransT \hfill (2021)  \cite{Chen2021} &53.90 & 80.12&83.88 & 51.20 & 78.10 & 80.00 \\
    KeepTrack  \hfill (2021) \cite{Mayer2021b} &51.44 & 74.20& 78.87&49.88 & 77.08&79.10  \\
    STARK  \hfill (2021) \cite{Yan2022} &50.99 &75.88 &77.16 &49.76 &77.12 &80.01  \\
    SimTrack \hfill (2022) \cite{chen2022backbone}&54.56 & 76.66& 78.10&50.66 &75.00 &78.88  \\
    MixFormer \hfill (2022)  \cite{cui2022mixformer} &59.88& 82.44& 85.14& 60.20 & 83.10 & 85.90 \\
    SwinTrack  \hfill (2022)\cite{lin2021swintrack} &53.99 & 79.18&81.22 & 52.40 & 78.80 & 80.30 \\
    RTS  \hfill (2022) \cite{Paul2022} &61.33&83.33& 89.70&63.20&81.67&88.67\\
    TOMP  \hfill (2022) \cite{mayer2022transforming} &50.21 & 78.90& 80.76& 51.10 & 75.20 & 81.80  \\
    OSTrack  \hfill (2022) \cite{ye2022ostrack} &60.10&80.91&88.43&59.10& 79.00&86.90 \\
    AiATrack  \hfill (2022) \cite{gao2022aiatrack} & 61.90& 81.32& 90.78&62.23 & 80.10& 85.60 \\
    DropTrack \hfill (2023)  \cite{dropmae2023} &59.88 & 81.00& 88.34&60.87 &80.90 & 85.10 \\
    GRM  \hfill (2023) \cite{gao2023generalized} &62.55& 79.88& 87.60&61.11&79.19&86.88 \\
    CiteTracker  \hfill (2023) \cite{li2023citetracker} & 60.98& 78.15& 85.60&59.19&78.14&85.10 \\
    ARTrack \hfill (2023)  \cite{Wei_2023_CVPR} &65.90 & 81.16& 87.60&64.12&79.10&84.89\\
    \hline
   \multirow{3}{*}{}BofN Meta-tracker & \multicolumn{3}{c|}{VOT2021 \cite{Kristan_2021_ICCV}} & \multicolumn{3}{c}{VOT2022 \cite{Kristan_2022_ICCV}} \\
   \cline{2-7}
    Video-Level&\underline{69.13}&\underline{86.70}&\underline{91.40}&\underline{67.88}&\underline{87.60}&\underline{90.87}  \\
    Frame-Level&\textbf{72.33}&\textbf{88.20}&\textbf{93.70}&\textbf{70.98}&\textbf{89.50}&\textbf{91.87} \\
    \hline
    \end{tabular}
    \label{table5}
\end{table*}

\subsection{BofN meta-tracker Performance Evaluation}
Table \ref{table3} shows the performance comparison of the proposed BofN meta-tracker on four datasets using different architectures employed during the full-finetuning of the proposed TP$^{2}$N (see Tables \ref{table1} $\&$ \ref{table2}).
Two different experimental settings are followed including frame-level evaluation and video-level evaluation.
In video-level evaluation, the best tracker is selected using initial frames and target location and then the same best tracker is used throughout that video sequence.
In frame-level evaluation, the best tracker is selected after every five video frames.
Overall, we observe that the frame-level evaluation obtains up to 4.17$\%$ better performance than the video-level evaluation.
This is because, in frame-level evaluation, the best tracker may switch depending on the scene and target object characteristics.
In all experiments, the ViT-S DINO has obtained the best performance compared to other methods.

\subsection{Qualitative Results}
To evaluate the performance of the proposed BofN tracker, we present visual results on key frames of 12 challenging sequences selected from the LaSOT and VOT2022 datasets as shown in Fig. \ref{fig_visual}.
The bounding boxes of the target objects are overlaid on the input images and the qualitative results comparisons are shown with  existing trackers. Overall, BofN tracker has performed much better
than the compared trackers in all sequences, which can be
attributed to the tracking performance prediction network 
in the proposed algorithm.

\subsection{Quantitative Evaluations: Performance Comparisons of BofN with SOTA Trackers}
Table \ref{table4} shows the performance comparison of the BofN meta-tracker with existing SOTA trackers on LaSOT, TrackingNet, and GOT-10K datasets.

Overall, both the video-level and frame-level BofN meta-trackers consistently outperform the SOTA visual trackers by a significant margin on all three datasets.
As discussed in Table \ref{table3}, BofN meta-tracker frame-level performance is better than the video-level performance.
More specifically, on the LaSOT dataset, we obtained 8.60$\%$ improvements in terms of AUC over the existing best  ARTrack tracker.
On TrackingNet, the BofN meta-tracker frame-level obtained 6.50$\%$ improvements in terms of AUC measure over the existing best  ARTrack tracker.
Similarly, on the GOT-10K dataset, the BofN meta-tracker frame-level obtained 5.90$\%$ improvements in terms of AO metric over the existing best  RTS tracker.

Table \ref{table5} shows the performance comparison of the proposed BofN meta-tracker with the 17 best SOTA visual trackers on VOT2021 and VOT2022 datasets.
Similar to Table \ref{table4}, we observed the same trend where BofN meta-tracker frame-level evaluation is better than the BofN meta-tracker video-level evaluation.
On the VOT2021 dataset, the BofN meta-tracker frame-level obtains 6.43$\%$ improvements in terms of EAO measure over the existing best-performing ARTrack tracker.
On the VOT2022 dataset, the BofN meta-tracker frame-level obtains 6.86$\%$ improvements in terms of EAO measure over the existing best-performing ARTrack tracker.

Table \ref{table_additional} shows the results on four additional datasets including OTB100, UAV123, VOT2019, and WebUAV-3M. The AUC performance of the proposed BofN meta-tracker is compared with the best SOTA tracker on these datasets.
Overall, these experiments demonstrate the effectiveness of the BofN meta-tracker both at the frame level and at the video-level evaluation protocols.

\begin{table}[t!]
\caption{BofN results (both video and frame level prediction) and the best-published tracker results on UAV123, OTB100, VOT2019, and WebUAV-3M datasets. BofN is superior to SOTA on all datasets, even for video-level prediction.}
\begin{center}
\makebox[\linewidth]{
\scalebox{0.78}{
\begin{tabu}{|c|c|c|c|}
\tabucline[0.5pt]{-}
Datasets&BofN - video &BofN - frame&Best SOTA: \hfill Performance\\\tabucline[0.5pt]{-}
UAV123 \textcolor{black}{\cite{mueller2016benchmark}}&\textbf{74.6}&\textbf{76.8}&ARTrack \textcolor{black}{\cite{Wei_2023_CVPR}}: \hfill 71.2\\\tabucline[0.5pt]{-}
OTB100 \textcolor{black}{\cite{otb100}}&\textbf{73.8}&\textbf{77.1}&ToMP \textcolor{black}{\cite{mayer2022transforming}}: \hfill 70.1\\\tabucline[0.5pt]{-}
VOT2019 \textcolor{black}{\cite{VOT2019}}&\textbf{77.2}&\textbf{79.6}&OSTrack \textcolor{black}{\cite{ye2022ostrack}}: \hfill 75.4\\\tabucline[0.5pt]{-}
WebUAV-3M \textcolor{black}{\cite{webuav_3M}}&\textbf{72.5}&\textbf{76.3}&DropTrack \textcolor{black}{\cite{dropmae2023}}: \hfill 69.4\\\tabucline[0.5pt]{-}
\end{tabu}
}}
\end{center}
\label{table_additional}
\end{table}

\begin{table}[t!]
\caption{BofN video-level meta-tracker tracking speed comparison in terms of fps with SOTA trackers.}
\begin{center}
\makebox[\linewidth]{
\scalebox{0.85}{
\begin{tabu}{|c||c|c|c|}
\tabucline[0.5pt]{-}
\textbf{Trackers}&LaSoT&VOT2022&TrackingNet\\\tabucline[0.5pt]{-}
BofN&68.91&78.56&65.54\\\tabucline[0.5pt]{-}
CiteTracker&80.20&82.31&78.44\\\tabucline[0.5pt]{-}
ARTrack&44.10&47.32&46.91\\\tabucline[0.5pt]{-}
GRM&43.44&42.54&45.35\\\tabucline[0.5pt]{-}
DropTrack&57.66&60.22&59.23\\\tabucline[0.5pt]{-}
\end{tabu}
}}
\end{center}
\label{tab_fps_compare}
\end{table}

\begin{table*}[t!]
\centering
\caption{\textbf{BofN meta-tracker} sensitivity both at the Frame and Video levels to size $N$ of the trackers pool. The performance is shown in terms of AUC for both LaSOT and TrackingNet, AO  for GOT-10K, and EAO for VOT2022. }
\makebox[\linewidth]{
\scalebox{0.98}{
\begin{tabular}{ c| c c| c c| c c| c c |c c |c c}
\hline
\textbf{Datasets} & \multicolumn{2}{|c|}{Bof3} & \multicolumn{2}{c|}{Bof6} & \multicolumn{2}{c|}{Bof9} & \multicolumn{2}{c}{Bof12}& \multicolumn{2}{|c}{Bof15}&\multicolumn{2}{|c}{Bof17}\\
& Video & Frame & Video & Frame & Video & Frame & Video & Frame & Video & Frame&Video & Frame\\
\hline
LaSOT&74.2&76.5&74.8&77.3&75.8&79.2&76.6&80.3&77.2&80.9&77.6&81.7\\
TrackingNet&86.1&88.6&86.3&89.1&86.6&89.5&87.3&90.2&87.8&91.6&88.9&92.1\\
GOT-10K&85.6&87.2&86.2&87.9&86.6&88.3&86.9&89.5&88.3&90.7&88.7&91.1\\
VOT2022&65.2&66.3&65.6&67.3&66.1&67.9&66.4&68.6&67.3&69.7&67.9&71.0\\
\hline
\end{tabular}
}}
\label{table6}
\end{table*}

\begin{table*}[t!]
    \centering
    \caption{BofN Meta-tracker. Tracker selected in the pool. }
    \begin{tabular}{ c|c|c|c|c|c|c}
    \hline
    \multirow{2}{*}{}Datasets & Bof3& Bof6&Bof9&Bof12&Bof15&Bof17 \\
    &Tracker& + & + & -- & -- &Tracker\\
    LaSOT& A B C &D E F & G H I& M N O & P Q & All\\
    TrackingNet& A D E & E F G & M K N & X Y X & U V & All\\
    GOT-10K & A D E & E F G & M K N & X Y X & U V & All\\
     VOT2022 & A D E & E F G & M K N & X Y X & U V & All\\
    \hline
    \end{tabular}
    \label{table7}
\end{table*}

\subsection{Ablation Studies}
\subsubsection{BofN Performance sensitivity to $N$ (size of the tracker's pool):} To evaluate the effect of $N$ on the performance of the BofN meta-tracker, we repeat the experiments with $N$=\{3, 6, 9, 12, 15, 17\} best trackers for each of the four datasets in both frame-level and video-level settings as shown Table \ref{table6}.
The SOTA trackers are arranged according to their performance on a particular dataset and the best $N$ trackers are selected in each experiment.
Table \ref{tab_sorting} shows the exact combination of the SOTA trackers for each dataset.
Overall, we observe an increase in the performance with the increasing size of the tracker pool from Bof3 to Bof17. 
A further increase in the pool size may further increase the performance of the BofN meta-tracker.

\begin{table*}[]
\centering
\caption{SOTA visual trackers in sorting order in terms of their performance on each dataset. Evaluated trackers: AiATrack (2022) \cite{gao2022aiatrack}, ARTrack (2023) \cite{Wei_2023_CVPR}, CiteTracker (2023) \cite{li2023citetracker}, DropTrack (2023) \cite{dropmae2023}, GRM (2023) \cite{gao2023generalized}, KeepTrack (2021) \cite{Mayer2021b}, MixFormer (2022) \cite{cui2022mixformer}, Ocean (2020) \cite{Ocean_2020_ECCV}, OSTrack (2022) \cite{ye2022ostrack}, RTS (2022) \cite{Paul2022}, SiamFC (2016) \cite{bertinetto2016fully}, SiamRPN++ (2018) \cite{Li2019a}, SimTrack (2022) \cite{chen2022backbone}, STARK (2021) \cite{Yan2022}, SwinTrack (2022) \cite{lin2021swintrack}, TOMP (2022) \cite{mayer2022transforming}, TransT (2021) \cite{Chen2021}.} 
\begin{tabular}{ |c|c|c|c|c|}
\hline
RANK & LaSOT \cite{8954084}& TrackingNet \cite{Muller_2018_ECCV} & GOT-10k \cite{8922619}& VOT2022 \cite{Kristan_2022_ICCV}  \\
\hline
A & ARTrack & ARTrack & RTS & ARTrack\\
B & DropTrack & CiteTracker & ARTrack & RTS \\ 
C & SwinTrack & DropTrack & DropTrack & AiATrack \\ 
D & OSTrack & GRM & CiteTracker & GRM \\ 
E & SimTrack & SwinTrack & OSTrack & DropTrack \\ 
F & MixFormer & OSTrack & GRM & MixFormer \\ 
G & GRM & MixFormer & SwinTrack & Ocean \\ 
H & CiteTracker & SimTrack & MixFormer & CiteTracker \\ 
I & RTS & AiATrack & TOMP & OSTrack \\
J & AiATrack & STARK & AiATrack & SwinTrack \\ 
K & TOMP & RTS & SimTrack & TransT \\ 
L & KeepTrack & TOMP & STARK & TOMP \\ 
M & STARK & TransT & TransT & SimTrack \\
N & TransT & KeepTrack & KeepTrack& KeepTrack \\ 
O & Ocean& SiamRPN++ & Ocean & STARK \\ 
P & SiamRPN++& Ocean & SianRPN++ & SiamRPN++\\ 
Q & SiamFC& SiamFC & SiamFC & SiamFC \\
\hline
\end{tabular}
\label{tab_sorting}
\end{table*}

\subsection{BofN meta-Tracker Overhead}
We compute the overhead of the TP$^2$N for the prediction of the BofN meta-tracker both at frame-level and video-level settings using $N$=17.
In video-level experiments, since only one evaluation is performed on each video, therefore, the overhead cost is 0.84s which is quite insignificant.

In the case of frame-level evaluation, the overhead cost depends on the number of times the best tracker is predicted in each video sequence.
If the time interval between two prediction is larger the computational overhead will be smaller.
In frame-level, the overall computational overhead will be 0.84 $\times n_{e}$, where $n_e$ is the number of evaluations.


Table \ref{tab_fps_compare} shows the tracking speed comparison with other SOTA methods on four datasets.
We observe that  BofN video-level meta-trakcer is significantly faster than the ARTrack, GRM, and DropTrack.
It is because the best performing trackers selected by BofN on these datasets are faster than these compared trackers.
We observe, that CiteTracker has a higher tracking speed while its AUC on LaSOT is 69.7$\%$ compared to 77.6$\%$ obtained by video-level BofN meta-tracker.
Thus, there is a tradeoff between computational cost and the performance.

\section{Conclusion \& Future Directions}
\label{sec:conclusion}
This work is based on the observation that SOTA visual tracker performance surprisingly strongly varies across different video attributes and datasets.
To exploit the phenomenon we propose the BofN meta-tracker that outperforms all state-of-the-art trackers by a very large margin. The core of the BofN meta-tracker is the Tracking Performance Prediction Network (TP$^2$N) that selects a tracker suitable for a given video sequence, using the first five frames only,  or for a set of frames within a temporal window.
The tracking performance prediction network is based on self-supervised learning architectures including MocoV2, SwAv, BT, and DINO. Experiments show that the DINO with ViT-S as a backbone performs best.
With this network, the BofN meta-tracker significantly outperforms the 17 SOTA trackers on each of the five datasets -- LaSOT, TrackingNet, GOT-10K, VOT2021, and VOT2022.
The proposed prediction paradigm is generic and can be easily extended to other computer vision tasks such as image classification, object detection, and segmentation, as well as other tracking methods.
As a side effect, the paper presents an extensive evaluation of competitive tracking methods on all commonly used benchmarks, following their protocols.
Besides its potential in applications, $BofN$ meta-tracker may be used for the generation of data for self-supervised learning.

\bibliographystyle{splncs04}
\bibliography{main}

\begin{thebibliography}{10}
\providecommand{\url}[1]{\texttt{#1}}
\providecommand{\urlprefix}{URL }
\providecommand{\doi}[1]{https://doi.org/#1}

\bibitem{beddiar2020vision}
Beddiar, D.R., Nini, B., Sabokrou, M., Hadid, A.: Vision-based human activity
  recognition: a survey. Multimedia Tools and Applications  \textbf{79}(41-42),
   30509--30555 (2020)

\bibitem{ben2022graph}
Ben-Haim, T., Raviv, T.R.: Graph neural network for cell tracking in microscopy
  videos. In: European Conference on Computer Vision. pp. 610--626. Springer
  (2022)

\bibitem{bertinetto2016fully}
Bertinetto, L., Valmadre, J., Henriques, J.F., Vedaldi, A., Torr, P.H.:
  Fully-convolutional siamese networks for object tracking. In: Computer
  Vision--ECCV 2016 Workshops: Amsterdam, The Netherlands, October 8-10 and
  15-16, 2016, Proceedings, Part II 14. pp. 850--865. Springer (2016)

\bibitem{Bhat_2018_ECCV}
Bhat, G., Johnander, J., Danelljan, M., Khan, F.S., Felsberg, M.: Unveiling the
  power of deep tracking. In: Proceedings of the European Conference on
  Computer Vision (ECCV) (September 2018)

\bibitem{10.1007/978-3-030-58452-8_13}
Carion, N., Massa, F., Synnaeve, G., Usunier, N., Kirillov, A., Zagoruyko, S.:
  End-to-end object detection with transformers. In: Vedaldi, A., Bischof, H.,
  Brox, T., Frahm, J.M. (eds.) Computer Vision -- ECCV 2020. pp. 213--229.
  Springer International Publishing, Cham (2020)

\bibitem{caron2020unsupervised}
Caron, M., Misra, I., Mairal, J., Goyal, P., Bojanowski, P., Joulin, A.:
  Unsupervised learning of visual features by contrasting cluster assignments.
  Advances in neural information processing systems  \textbf{33},  9912--9924
  (2020)

\bibitem{caron2021emerging}
Caron, M., Touvron, H., Misra, I., J{\'e}gou, H., Mairal, J., Bojanowski, P.,
  Joulin, A.: Emerging properties in self-supervised vision transformers. In:
  Proceedings of the IEEE/CVF international conference on computer vision. pp.
  9650--9660 (2021)

\bibitem{9354012}
Chang, X., Pan, H., Sun, W., Gao, H.: Yoltrack: Multitask learning based
  real-time multiobject tracking and segmentation for autonomous vehicles. IEEE
  Transactions on Neural Networks and Learning Systems  \textbf{32}(12),
  5323--5333 (2021). \doi{10.1109/TNNLS.2021.3056383}

\bibitem{chen2022backbone}
Chen, B., Li, P., Bai, L., Qiao, L., Shen, Q., Li, B., Gan, W., Wu, W., Ouyang,
  W.: Backbone is all your need: A simplified architecture for visual object
  tracking pp. 375--392 (2022)

\bibitem{chen2019multi}
Chen, B., Li, P., Sun, C., Wang, D., Yang, G., Lu, H.: Multi attention module
  for visual tracking. Pattern Recognition  \textbf{87},  80--93 (2019)

\bibitem{chen2020simple}
Chen, T., Kornblith, S., Norouzi, M., Hinton, G.: A simple framework for
  contrastive learning of visual representations. In: International conference
  on machine learning. pp. 1597--1607. PMLR (2020)

\bibitem{Chen2021}
Chen, X., Yan, B., Zhu, J., Wang, D., Yang, X., Lu, H.: Transformer tracking.
  In: Proceedings of the IEEE/CVF Conference on Computer Vision and Pattern
  Recognition (CVPR). pp. 8126--8135 (June 2021)

\bibitem{chen2020improved}
Chen, X., Fan, H., Girshick, R., He, K.: Improved baselines with momentum
  contrastive learning. arXiv preprint arXiv:2003.04297  (2020)

\bibitem{Choi_2017_CVPR}
Choi, J., Jin~Chang, H., Yun, S., Fischer, T., Demiris, Y., Young~Choi, J.:
  Attentional correlation filter network for adaptive visual tracking. In:
  Proceedings of the IEEE Conference on Computer Vision and Pattern Recognition
  (CVPR) (July 2017)

\bibitem{cui2022mixformer}
Cui, Y., Jiang, C., Wang, L., Wu, G.: Mixformer: End-to-end tracking with
  iterative mixed attention. In: Proceedings of the IEEE/CVF Conference on
  Computer Vision and Pattern Recognition. pp. 13608--13618 (2022)

\bibitem{Dai_2020_CVPR}
Dai, K., Zhang, Y., Wang, D., Li, J., Lu, H., Yang, X.: High-performance
  long-term tracking with meta-updater. In: IEEE/CVF Conference on Computer
  Vision and Pattern Recognition (CVPR) (June 2020)

\bibitem{7569092}
Danelljan, M., Häger, G., Khan, F.S., Felsberg, M.: Discriminative scale space
  tracking. IEEE Transactions on Pattern Analysis and Machine Intelligence
  \textbf{39}(8),  1561--1575 (2017). \doi{10.1109/TPAMI.2016.2609928}

\bibitem{dosovitskiy2020image}
Dosovitskiy, A., Beyer, L., Kolesnikov, A., Weissenborn, D., Zhai, X.,
  Unterthiner, T., Dehghani, M., Minderer, M., Heigold, G., Gelly, S., et~al.:
  An image is worth 16x16 words: Transformers for image recognition at scale.
  arXiv preprint arXiv:2010.11929  (2020)

\bibitem{dunnhofer2022combining}
Dunnhofer, M., Simonato, K., Micheloni, C.: Combining complementary trackers
  for enhanced long-term visual object tracking. Image and Vision Computing
  \textbf{122},  104448 (2022)

\bibitem{7428948}
Fan, C.T., Wang, Y.K., Huang, C.R.: Heterogeneous information fusion and
  visualization for a large-scale intelligent video surveillance system. IEEE
  Transactions on Systems, Man, and Cybernetics: Systems  \textbf{47}(4),
  593--604 (2017). \doi{10.1109/TSMC.2016.2531671}

\bibitem{8954084}
Fan, H., Lin, L., Yang, F., Chu, P., Deng, G., Yu, S., Bai, H., Xu, Y., Liao,
  C., Ling, H.: Lasot: A high-quality benchmark for large-scale single object
  tracking. In: 2019 IEEE/CVF Conference on Computer Vision and Pattern
  Recognition (CVPR). pp. 5369--5378 (2019). \doi{10.1109/CVPR.2019.00552}

\bibitem{8667882}
Fan, H., Ling, H.: Parallel tracking and verifying. IEEE Transactions on Image
  Processing  \textbf{28}(8),  4130--4144 (2019).
  \doi{10.1109/TIP.2019.2904789}

\bibitem{Fan_2019_CVPR}
Fan, H., Ling, H.: Siamese cascaded region proposal networks for real-time
  visual tracking. In: Proceedings of the IEEE/CVF Conference on Computer
  Vision and Pattern Recognition (CVPR) (June 2019)

\bibitem{gao2022aiatrack}
Gao, S., Zhou, C., Ma, C., Wang, X., Yuan, J.: Aiatrack: Attention in attention
  for transformer visual tracking. In: European Conference on Computer Vision.
  pp. 146--164. Springer (2022)

\bibitem{gao2023generalized}
Gao, S., Zhou, C., Zhang, J.: Generalized relation modeling for transformer
  tracking. In: Proceedings of the IEEE/CVF Conference on Computer Vision and
  Pattern Recognition. pp. 18686--18695 (2023)

\bibitem{graham2019simultaneous}
Graham, S., Vu, Q., Raza, S., Azam, A., Tsang, Y., Kwak, J., Hover-Net, N.R.:
  Simultaneous segmentation and classification of nuclei in multi-tissue
  histology images., 2019, 58. DOI: https://doi. org/10.1016/j. media p. 101563
  (2019)

\bibitem{He_2018_CVPR}
He, A., Luo, C., Tian, X., Zeng, W.: A twofold siamese network for real-time
  object tracking. In: Proceedings of the IEEE Conference on Computer Vision
  and Pattern Recognition (CVPR) (June 2018)

\bibitem{8922619}
Huang, L., Zhao, X., Huang, K.: Got-10k: A large high-diversity benchmark for
  generic object tracking in the wild. IEEE Transactions on Pattern Analysis
  and Machine Intelligence  \textbf{43}(5),  1562--1577 (2021).
  \doi{10.1109/TPAMI.2019.2957464}

\bibitem{9913708}
Javed, S., Danelljan, M., Khan, F.S., Khan, M.H., Felsberg, M., Matas, J.:
  Visual object tracking with discriminative filters and siamese networks: A
  survey and outlook. IEEE Transactions on Pattern Analysis and Machine
  Intelligence  \textbf{45}(5),  6552--6574 (2023).
  \doi{10.1109/TPAMI.2022.3212594}

\bibitem{9666461}
Jiao, L., Wang, D., Bai, Y., Chen, P., Liu, F.: Deep learning in visual
  tracking: A review. IEEE Transactions on Neural Networks and Learning Systems
   \textbf{34}(9),  5497--5516 (2023). \doi{10.1109/TNNLS.2021.3136907}

\bibitem{kristan2022tenth}
Kristan, M., Leonardis, A., Matas, J., Felsberg, M., Pflugfelder, R.,
  K{\"a}m{\"a}r{\"a}inen, J.K., Chang, H.J., Danelljan, M., Zajc, L.{\v{C}}.,
  Luke{\v{z}}i{\v{c}}, A., et~al.: The tenth visual object tracking vot2022
  challenge results. In: European Conference on Computer Vision. pp. 431--460.
  Springer (2022)

\bibitem{VOT2019}
Kristan, M., Matas, J., Leonardis, A., Felsberg, M., Pflugfelder, R.,
  Kamarainen, J.K., ˇCehovin~Zajc, L., Drbohlav, O., Lukezic, A., Berg, A.,
  et~al.: The seventh visual object tracking vot2019 challenge results. In:
  {IEEE ICCVW} (2019)

\bibitem{Kristan_2021_ICCV}
Kristan, M., Matas, J., Leonardis, A., Felsberg, M., Pflugfelder, R.,
  K\"am\"ar\"ainen, J.K., Chang, H.J., Danelljan, M., Cehovin, L.,
  Luke\v{z}i\v{c}, A., Drbohlav, O., K\"apyl\"a, J., H\"ager, G., Yan, S.,
  Yang, J., Zhang, Z., Fern\'andez, G.: The ninth visual object tracking
  vot2021 challenge results. In: Proceedings of the IEEE/CVF International
  Conference on Computer Vision (ICCV) Workshops. pp. 2711--2738 (October 2021)

\bibitem{Kristan_2022_ICCV}
Kristan, M., Matas, J., Leonardis, A., Felsberg, M., Pflugfelder, R.,
  K\"am\"ar\"ainen, J.K., Chang, H.J., Danelljan, M., Cehovin, L.,
  Luke\v{z}i\v{c}, A., Drbohlav, O., K\"apyl\"a, J., H\"ager, G., Yan, S.,
  Yang, J., Zhang, Z., Fern\'andez, G.: The tenth visual object tracking
  vot2022 challenge results. In: Karlinsky, L., Michaeli, T., Nishino, K.
  (eds.) Computer Vision -- ECCV 2022 Workshops. pp. 431--460. Springer Nature
  Switzerland, Cham (2023)

\bibitem{LEANG2018459}
Leang, I., Herbin, S., Girard, B., Droulez, J.: On-line fusion of trackers for
  single-object tracking. Pattern Recognition  \textbf{74},  459--473 (2018).
  \doi{https://doi.org/10.1016/j.patcog.2017.09.026}

\bibitem{Li_2019_CVPR}
Li, B., Wu, W., Wang, Q., Zhang, F., Xing, J., Yan, J.: Siamrpn++: Evolution of
  siamese visual tracking with very deep networks. In: Proceedings of the
  IEEE/CVF Conference on Computer Vision and Pattern Recognition (CVPR) (June
  2019)

\bibitem{Li2019a}
Li, B., Wu, W., Wang, Q., Zhang, F., Xing, J., Yan, J.: Siamrpn++: Evolution of
  siamese visual tracking with very deep networks pp. 4282--4291 (2019)

\bibitem{9018389}
Li, H., Wu, X.J., Kittler, J.: Mdlatlrr: A novel decomposition method for
  infrared and visible image fusion. IEEE Transactions on Image Processing
  \textbf{29},  4733--4746 (2020). \doi{10.1109/TIP.2020.2975984}

\bibitem{li2023citetracker}
Li, X., Huang, Y., He, Z., Wang, Y., Lu, H., Yang, M.H.: Citetracker:
  Correlating image and text for visual tracking. In: Proceedings of the
  IEEE/CVF International Conference on Computer Vision. pp. 9974--9983 (2023)

\bibitem{lin2021swintrack}
Lin, L., Fan, H., Zhang, Z., Xu, Y., Ling, H.: Swintrack: A simple and strong
  baseline for transformer tracking (2022)

\bibitem{Lukezic_2017_CVPR}
Lukezic, A., Vojir, T., Cehovin~Zajc, L., Matas, J., Kristan, M.:
  Discriminative correlation filter with channel and spatial reliability. In:
  Proceedings of the IEEE Conference on Computer Vision and Pattern Recognition
  (CVPR) (July 2017)

\bibitem{Ma_2015_ICCV}
Ma, C., Huang, J.B., Yang, X., Yang, M.H.: Hierarchical convolutional features
  for visual tracking. In: Proceedings of the IEEE International Conference on
  Computer Vision (ICCV) (December 2015)

\bibitem{martinez2014animal}
Martinez-Martin, E., del Pobil, A.P.: Animal social behaviour: A visual
  analysis. In: International Conference on Simulation of Adaptive Behavior.
  pp. 320--327. Springer (2014)

\bibitem{mayer2022transforming}
Mayer, C., Danelljan, M., Bhat, G., Paul, M., Paudel, D.P., Yu, F., Van~Gool,
  L.: Transforming model prediction for tracking. In: Computer Vision and
  Pattern Recognition (2022)

\bibitem{Mayer2021b}
Mayer, C., Danelljan, M., Paudel, D.P., Van~Gool, L.: Learning target candidate
  association to keep track of what not to track. In: Proceedings of the
  IEEE/CVF International Conference on Computer Vision (ICCV). pp. 13444--13454
  (October 2021)

\bibitem{10.1007/978-3-319-46448-0_27}
Mueller, M., Smith, N., Ghanem, B.: A benchmark and simulator for uav tracking.
  In: Leibe, B., Matas, J., Sebe, N., Welling, M. (eds.) Computer Vision --
  ECCV 2016. pp. 445--461. Springer International Publishing, Cham (2016)

\bibitem{mueller2016benchmark}
Mueller, M., Smith, N., Ghanem, B.: A benchmark and simulator for uav tracking.
  In: {ECCV}. Springer (2016)

\bibitem{Muller_2018_ECCV}
Muller, M., Bibi, A., Giancola, S., Alsubaihi, S., Ghanem, B.: Trackingnet: A
  large-scale dataset and benchmark for object tracking in the wild. In:
  Proceedings of the European Conference on Computer Vision (ECCV) (September
  2018)

\bibitem{nam2016mdnet}
Nam, H., Han, B.: Learning multi-domain convolutional neural networks for
  visual tracking. In: The IEEE Conference on Computer Vision and Pattern
  Recognition (CVPR) (June 2016)

\bibitem{10.1007/978-3-031-20047-2_33}
Paul, M., Danelljan, M., Mayer, C., Van~Gool, L.: Robust visual tracking by
  segmentation. In: Avidan, S., Brostow, G., Ciss{\'e}, M., Farinella, G.M.,
  Hassner, T. (eds.) Computer Vision -- ECCV 2022. pp. 571--588. Springer
  Nature Switzerland, Cham (2022)

\bibitem{Paul2022}
Paul, M., Danelljan, M., Mayer, C., Van~Gool, L.: Robust visual tracking by
  segmentation. In: European Conference on Computer Vision. pp. 571--588.
  Springer (2022)

\bibitem{Qi_2016_CVPR}
Qi, Y., Zhang, S., Qin, L., Yao, H., Huang, Q., Lim, J., Yang, M.H.: Hedged
  deep tracking. In: Proceedings of the IEEE Conference on Computer Vision and
  Pattern Recognition (CVPR) (June 2016)

\bibitem{tang2023exploring}
Tang, Z., Xu, T., Li, H., Wu, X.J., Zhu, X., Kittler, J.: Exploring fusion
  strategies for accurate rgbt visual object tracking. Information Fusion p.
  101881 (2023)

\bibitem{Tao_2016_CVPR}
Tao, R., Gavves, E., Smeulders, A.W.: Siamese instance search for tracking. In:
  Proceedings of the IEEE Conference on Computer Vision and Pattern Recognition
  (CVPR) (June 2016)

\bibitem{thangavel2023transformers}
Thangavel, J., Kokul, T., Ramanan, A., Fernando, S.: Transformers in single
  object tracking: An experimental survey. arXiv preprint arXiv:2302.11867
  (2023)

\bibitem{tian2021divide}
Tian, Y., Henaff, O.J., van~den Oord, A.: Divide and contrast: Self-supervised
  learning from uncurated data. In: Proceedings of the IEEE/CVF International
  Conference on Computer Vision. pp. 10063--10074 (2021)

\bibitem{Valmadre_2017_CVPR}
Valmadre, J., Bertinetto, L., Henriques, J., Vedaldi, A., Torr, P.H.S.:
  End-to-end representation learning for correlation filter based tracking. In:
  Proceedings of the IEEE Conference on Computer Vision and Pattern Recognition
  (CVPR) (July 2017)

\bibitem{8578607}
Wang, N., Zhou, W., Tian, Q., Hong, R., Wang, M., Li, H.: Multi-cue correlation
  filters for robust visual tracking. In: 2018 IEEE/CVF Conference on Computer
  Vision and Pattern Recognition. pp. 4844--4853 (2018).
  \doi{10.1109/CVPR.2018.00509}

\bibitem{9316980}
Wang, R., Zhang, X., Fang, Y., Li, B.: Virtual-goal-guided rrt for visual
  servoing of mobile robots with fov constraint. IEEE Transactions on Systems,
  Man, and Cybernetics: Systems  \textbf{52}(4),  2073--2083 (2022).
  \doi{10.1109/TSMC.2020.3044347}

\bibitem{Wei_2023_CVPR}
Wei, X., Bai, Y., Zheng, Y., Shi, D., Gong, Y.: Autoregressive visual tracking.
  In: Proceedings of the IEEE/CVF Conference on Computer Vision and Pattern
  Recognition (CVPR). pp. 9697--9706 (June 2023)

\bibitem{585893}
Wolpert, D., Macready, W.: No free lunch theorems for optimization. IEEE
  Transactions on Evolutionary Computation  \textbf{1}(1),  67--82 (1997).
  \doi{10.1109/4235.585893}

\bibitem{dropmae2023}
Wu, Q., Yang, T., Liu, Z., Wu, B., Shan, Y., Chan, A.B.: Dropmae: Masked
  autoencoders with spatial-attention dropout for tracking tasks. In:
  Proceedings of the IEEE/CVF Conference on Computer Vision and Pattern
  Recognition. pp. 14561--14571 (2023)

\bibitem{otb100}
Wu, Y., Lim, J., Yang, M.H.: Object tracking benchmark. {IEEE TPAMI}
  \textbf{37}(9),  1834--1848 (2015)

\bibitem{Xu_2019_ICCV}
Xu, T., Feng, Z.H., Wu, X.J., Kittler, J.: Joint group feature selection and
  discriminative filter learning for robust visual object tracking. In:
  Proceedings of the IEEE/CVF International Conference on Computer Vision
  (ICCV) (October 2019)

\bibitem{Yan2022}
Yan, B., Peng, H., Fu, J., Wang, D., Lu, H.: Learning spatio-temporal
  transformer for visual tracking. In: Proceedings of the IEEE/CVF
  international conference on computer vision. pp. 10448--10457 (2021)

\bibitem{ye2022ostrack}
Ye, B., Chang, H., Ma, B., Shan, S., Chen, X.: Joint feature learning and
  relation modeling for tracking: A one-stream framework. In: European
  Conference on Computer Vision. pp. 341--357. Springer (2022)

\bibitem{you2017large}
You, Y., Gitman, I., Ginsburg, B.: Large batch training of convolutional
  networks. arXiv preprint arXiv:1708.03888  (2017)

\bibitem{zbontar2021barlow}
Zbontar, J., Jing, L., Misra, I., LeCun, Y., Deny, S.: Barlow twins:
  Self-supervised learning via redundancy reduction. In: International
  Conference on Machine Learning. pp. 12310--12320. PMLR (2021)

\bibitem{webuav_3M}
Zhang, C., Huang, G., Liu, L., Huang, S., Yang, Y., Wan, X., Ge, S., Tao, D.:
  Webuav-3m: A benchmark for unveiling the power of million-scale deep uav
  tracking. IEEE Transactions on Pattern Analysis and Machine Intelligence
  \textbf{45}(7),  9186--9205 (2023). \doi{10.1109/TPAMI.2022.3232854}

\bibitem{10.1007/978-3-319-10599-4_13}
Zhang, J., Ma, S., Sclaroff, S.: Meem: Robust tracking via multiple experts
  using entropy minimization. In: Fleet, D., Pajdla, T., Schiele, B.,
  Tuytelaars, T. (eds.) Computer Vision -- ECCV 2014. pp. 188--203. Springer
  International Publishing, Cham (2014)

\bibitem{Ocean_2020_ECCV}
Zhang, Z., Peng, H., Fu, J., Li, B., Hu, W.: Ocean: Object-aware anchor-free
  tracking. In: Computer Vision--ECCV 2020: 16th European Conference, Glasgow,
  UK, August 23--28, 2020, Proceedings, Part XXI 16. pp. 771--787. Springer
  (2020)

\bibitem{ZHAO2021107679}
Zhao, S., Xu, T., Wu, X.J., Zhu, X.F.: Adaptive feature fusion for visual
  object tracking. Pattern Recognition  \textbf{111},  107679 (2021).
  \doi{https://doi.org/10.1016/j.patcog.2020.107679}

\end{thebibliography}

\vspace{-15mm}
\begin{IEEEbiography}
	[{\includegraphics[width=1in,height=2.0in,clip,keepaspectratio]{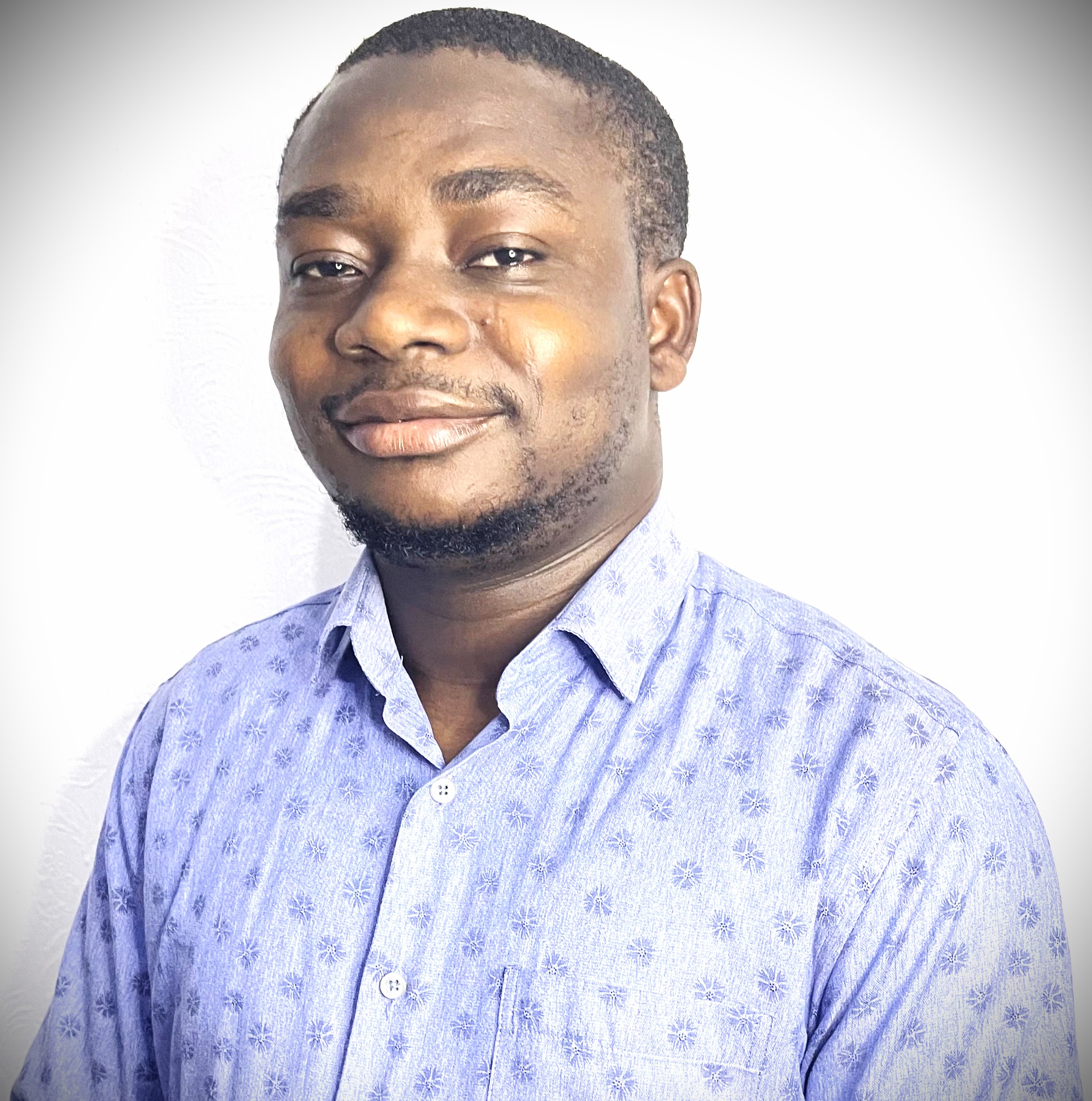}}]{Basit Alawode}
received his BSc degree in electrical engineering from the Obafemi Awolowo University, Nigeria, in 2013. 
He further obtained his MSc degree in electrical engineering at the King Fahd University of Petroleum and Minerals, Saudi Arabia in 2020. 
He is currently pursuing his Ph.D. degree in computer science and engineering degree at Khalifa University of Science and Technology, UAE.
His research interests include visual object tracking and computer vision.
\end{IEEEbiography}
\vspace{-15mm}
\begin{IEEEbiography}
	[{\includegraphics[width=1in,height=2.0in,clip,keepaspectratio]{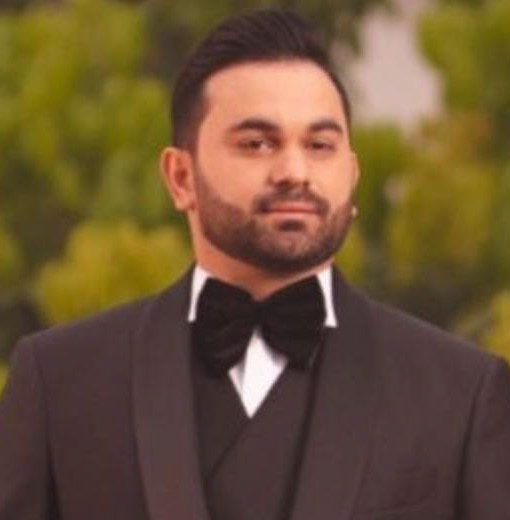}}]{Sajid Javed}
is a faculty member at Khalifa University (KU), UAE. 
Prior to that, he was a research fellow at KU from 2019 to 2021 and at University of Warwick, U.K, from 2017-2018. 
He received his B.Sc. degree in computer science from University of Hertfordshire, U.K, in 2010. 
He completed his combined Master’s and Ph.D. degree in computer science from Kyungpook National University, Republic of Korea, in 2017.
\end{IEEEbiography}

\begin{IEEEbiography}
[{\includegraphics[width=1in,height=1.35in,clip,keepaspectratio]{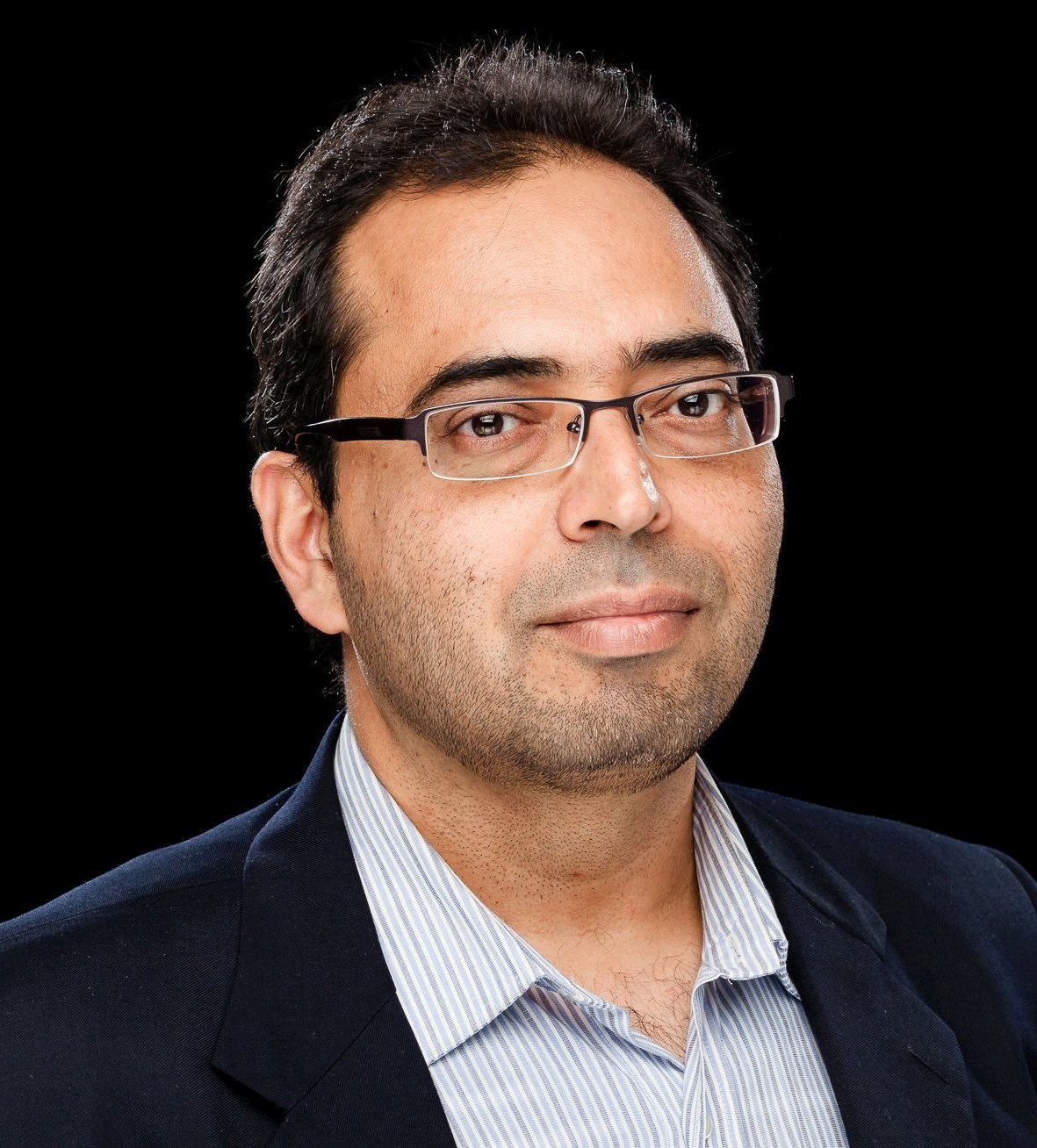}}]{Arif Mahmood}
is currently a professor at the Department of Computer Science at ITU, Pakistan, and the director of the computer vision lab.
He received his M.Sc. and Ph.D. degree in computer science from LUMS, Pakistan, in 2003 and 2011.
He has also worked as a Research Assistant Professor with the School of Mathematics and Statistics (SMS), University of Western Australia (UWA).
His research interests are face recognition, object classification, human-object interaction detection, and abnormal event detection. 
\end{IEEEbiography}

\begin{IEEEbiography}
	[{\includegraphics[width=1in,height=1.20in,clip,keepaspectratio]{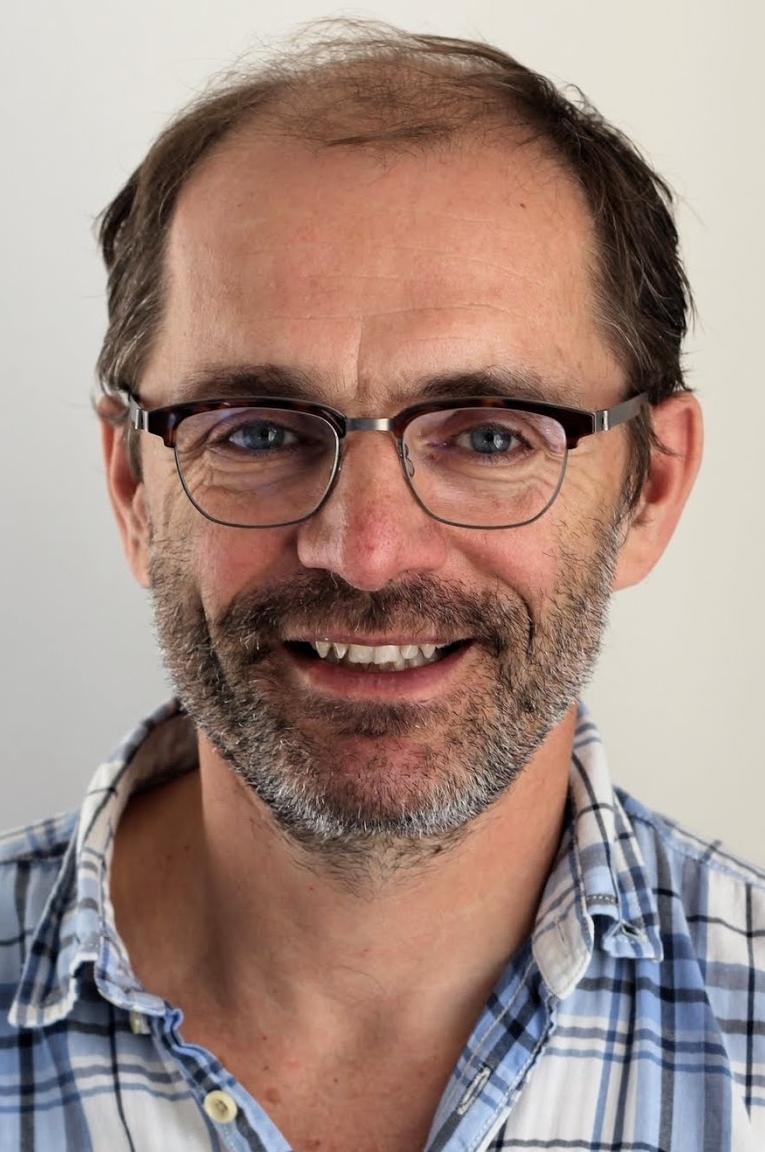}}]{Jiri Matas}
is a full professor at Czech Technical University in Prague.
He holds a PhD degree from the University of Surrey, UK (1995).
He has published more than 250 papers in refereed journals and conferences. 
His publications have about 53000 citations registered in Google Scholar; his h-index is 85.
J. Matas has served in various roles at major international computer vision conferences.
He is an EiC of IJCV and was an AE of IEEE T. PAMI. 
His research interests include visual tracking, object recognition, image matching and retrieval, sequential pattern recognition, and RANSAC- type optimization methods.
\end{IEEEbiography}

\end{document}